\documentclass[sn-basic,pdflatex]{sn-jnl}

\usepackage{graphicx}%
\usepackage{multirow}%
\usepackage{amsmath,amssymb,amsfonts}%
\usepackage{amsthm}%
\usepackage{mathrsfs}%
\usepackage[title]{appendix}%
\usepackage{xcolor}%
\usepackage{textcomp}%
\usepackage{manyfoot}%
\usepackage{booktabs}%
\usepackage{algorithm}%
\usepackage{algorithmicx}%
\usepackage{algpseudocode}%
\usepackage{listings}%

\theoremstyle{thmstyleone}%
\theoremstyle{thmstyletwo}%

\theoremstyle{thmstylethree}%
\newtheorem{definition}{Definition}%

\renewcommand*{\vec}[1]{\mathbf{#1}}
\newcommand*{\set}[1]{\{#1\}}  
\newcommand*{\p}[1]{{p(\vec{#1})}}       
\newcommand*{\pp}[1]{{p_\theta({#1})}} 
\newcommand*{\q}[1]{{q(\vec{#1})}}
\newcommand*{\qq}[1]{{q_\phi({#1})}} 
   
\newcommand{\n}{\mathbf n} 
\newcommand{\s}{\mathbf s} 
 
\newcommand{\x}{\mathbf x}   
\newcommand{\y}{\mathbf y}   
\newcommand{\z}{\mathbf z}   
\newcommand{\E}{\mathbb E} 

\renewcommand*{\H}[1]{H(\vec{#1})}    

\newcommand*{\I}[1]{I(\vec{#1})}      
      
\newcommand{\D}{D_\text{KL}}

\usepackage{listings} 
\usepackage{amsmath,amsfonts,amssymb}
\usepackage{subcaption}
\usepackage{xfrac}
\usepackage{booktabs, multirow, siunitx}
\usepackage[section]{placeins}

\begin{document}

\title[Trustworthy Representation Learning via Information Funnels and Bottlenecks]{Trustworthy Representation Learning via Information Funnels and Bottlenecks}


\author*[1,2,3]{\fnm{João} \sur{Machado de Freitas}}\email{jfreitas@ieee.org}

\author[2,3]{\fnm{Bernhard} \sur{C. Geiger}}\email{bgeiger@ieee.org}

\affil[1]{\orgdiv{Christian Doppler Laboratory for Dependable Intelligent Systems in Harsh Environments}, \orgname{Graz University of Technology}, \orgaddress{\street{Inffeldgasse 16c}, \city{Graz}, \postcode{8010}, \state{Steiermark}, \country{Austria}}}

\affil[2]{\orgdiv{Signal Processing and Speech Communication Laboratory}, \orgname{Graz University of Technology}, \orgaddress{\street{Inffeldgasse 16c}, \city{Graz}, \postcode{8010}, \state{Steiermark}, \country{Austria}}}

\affil[3]{\orgname{Know Center Research GmbH}, \orgaddress{\street{Sandgasse 34/2}, \city{Graz}, \postcode{8010}, \state{Steiermark}, \country{Austria}}}



\abstract{Ensuring trustworthiness in machine learning --- by balancing utility, fairness, and privacy --- remains a critical challenge, particularly in representation learning. In this work, we investigate a family of closely related information-theoretic objectives, including information funnels and bottlenecks, designed to extract invariant representations from data. We introduce the Conditional Privacy Funnel with Side-information (CPFSI), a novel formulation within this family, applicable in both fully and semi-supervised settings. Given the intractability of these objectives, we derive neural-network-based approximations via amortized variational inference. We systematically analyze the trade-offs between utility, invariance, and representation fidelity, offering new insights into the Pareto frontiers of these methods. Our results demonstrate that CPFSI effectively balances these competing objectives and frequently outperforms existing approaches. Furthermore, we show that by intervening on sensitive attributes in CPFSI's predictive posterior enhances fairness while maintaining predictive performance. Finally, we focus on the real-world applicability of these approaches, particularly for learning robust and fair representations from tabular datasets in data scarce-environments --- a modality where these methods are often especially relevant.}

\maketitle

\section{Introduction}
\label{sec:introduction}

Real-world decision-making applications raise the need to filter out the influence of specific attributes to meet legislative standards or to address ethical concerns tied to machine learning (ML) deployment. While traditionally ML research has overlooked potential biases in the training data and prioritized model performance, there has been a shift in focus in recent times. A growing emphasis is now being placed on fair, accountable, and trustworthy research, and the incorporation of objectives beyond simple predictive performance.
Further, there is a notable lack of research probing how ML can satisfy these often-conflicting requirements in real-world application scenarios (such as small tabular datasets) and, an over-reliance on toy examples from computer vision, like ColoredMNIST or CelebA. 
While there has recently been work by the deep learning community targeting tabular data, such as TabNet~\citep{arik2021tabnet}, deriving useful, invariant representations from unlabeled or partially labeled data remains a significant challenge.
Consequently, there is a clear demand for more research to overcome these barriers and advance the development of more robust and reliable ML systems.

Both private representation learning and fair representation learning are subfields of ML that emphasize the protection of sensitive data features; see~\citep{mehrabi2021survey,huang2020survey} and Section~\ref{sec:relatedwork}.
Private representation learning aims to generate representations that are useful but make it difficult for an adversary to recover private or sensitive data attributes (refer to the discussion for a clarification regarding the utilization of the term “privacy”).
Fair representation learning seeks to devise representations that are beneficial for a task execution, while ensuring the non-disclosure of information that could lead to discriminatory or unfair decisions. 
The common thread binding these two approaches is the dual goal of preserving utility and a notion of invariance in data representation in relation to a specified context variable, such as gender or age, that needs to be obscured or protected.
Managing these trade-offs is challenging, but invariant representation learning promises to reduce the influence of known confounders, contribute to the robustness of ML systems, and improve out-of-domain generalization.

In this work, we investigate a family of information funnel --- and bottleneck --- objectives and propose a novel member: the Conditional Privacy Funnel with Side-Information (CPFSI, Section~\ref{sec:CPFSI}). This model aims to produce private/fair representations that are faithful to the input and ensure robust predictive performance on a designated downstream task.
Concretely, CPFSI encodes the input covariates $\x$ to a learned representation $\z$, which is approximately independent of a sensitive attribute $\s$, but allows reconstructing $\x$ and predicting the target variable $\y$.

Unlike previous work~\citep{makhdoumi2014information, rodriguez2021variational}, CPFSI does not focus solely on creating representations that remain useful for unspecified downstream tasks. Instead, our approach builds upon the existing work in information-theoretic objectives for unsupervised representation learning, and uses side-information about a task. This new feature is useful to adapt our method to weak-supervision and is illustrated with experiments in the semi-supervised setting, where only a small amount of labeled data per class is available. 
This was motivated by the fundamental difficulty in learning meaningful representations without some sort of supervision, as shown in~\citep{locatello2019challenging,locatello2020disentangling}. 

The CPFSI is a missing link in the family of objectives described in Section~\ref{sec:relatedwork}. In particular, we extend previous work by providing more detailed results and characterizing empirically the entire family of information-theoretic objectives in terms of representation's fairness and privacy, and fair classification in fully and semi-supervised settings. In this work, the experiments are mainly aimed to extend the existing literature by painting a clearer picture about the trade-offs involved with these methods and their real-world practicality for small amounts of tabular data. 

To ensure a fair comparison between methods, the CPFSI reuses variational bounds and approximations similar to those that have been employed for other information-theoretic objectives (Section~\ref{sec:CPFSI:variational}).
We are aware that the technique chosen to approximate these objectives --- in this case deep amortized variational inference --- has a profound effect on the quality of the representations. Indeed, it has been shown that different variational approximations to otherwise equivalent information-theoretic objectives can result in substantially different behavior of the trained ML system~\citep{geiger2020comparison}. 
While we believe to be the case also for CPFSI, we defer this investigation to future work. 

\textbf{Notation.} We denote random variables as bold, e.g., $\x$ and realizations as standard letters, e.g., $x$. Distributions and variational approximations thereof are denoted with $p$ and $q$, respectively. We write $\I{\cdot;\cdot}$ for mutual information.

\section{Related Work}
\label{sec:relatedwork}

One of the most prominent methods for learning representations for a downstream task is the information bottleneck (IB) method~\citep{tishby2001information}. Based on rate-distortion theory, it aims for a mapping $\p{z|x}$ that maximizes the data compression $\I{z;x}$ while simultaneously ensuring that the representation $\z$ is informative for the downstream task, i.e., informative about $\y$. Its Lagrangian formulation thus aims to find $\p{z|x}$ by minimizing $\mathcal L_\text{IB}:= \I{z;x} - \rho \I{z;y}$, where $\rho$ trades between compressing and preserving meaningful information. \citet{alemi2017deep} provided a variational bound to approximate this objective.

Subsequently to~\citep{tishby2001information}, the IB method was extended by~\citet{chechik2002extracting} to learn representations $\z$ that are informative about $\y$ but exclude information about another variable $\s$. In other words, the IB with side-information (IBSI) --- also known as discriminative IB --- restricts the amount of irrelevant information $\I{z;s}$, ergo, learns representations $\z$ invariant to $\s$ for a particular task $\y$. Assuming the conditional independence $\y \bot \s | \x$, the Lagrangian formulation of the IBSI problem seeks a mapping $\p{z|x}$ minimizing
\begin{align}
\label{eq:IBSI}
    \mathcal L_\text{IBSI}&= \I{z;x} + \gamma \I{s;z} - \lambda \I{y;z}\\
    &= \I{z;x} - \alpha \I{z;x|s} - \beta \I{y;z},
\end{align}
where the Lagrange multipliers $\alpha=\gamma/(1+\gamma)\in[0,1)$ and $\beta=\lambda/(1+\gamma)\ge 0$
determine the trade-offs between compression and information extraction, and between loss of information about $\s$ and preservation of information about $\y$, respectively. Apparently unaware of~\citep{chechik2002extracting},~\citet[Section~2.2]{moyer2018invariant} proposed a deep variational approximation to the objective in Equation~\eqref{eq:IBSI}. 
Furthermore, the Complexity-Leakage-Utility Bottleneck (CLUB) objective has been proposed in~\citep{CLUB}, where the authors also aimed for a compressed representation that is informative about a task and that does not leak any sensitive information. The CLUB Lagrangian is equivalent to the one of IBSI, but was implemented using a variational approach that requires adversarial training.

An alternative (conditional) approach to IBSI is the Conditional Fairness Bottleneck (CFB) proposed in~\citep{rodriguez2021variational}, which also seeks compressed representations for invariant representation learning by minimizing
\begin{align}
    \mathcal L_\text{CFB} :&= \I{s;z} + \I{z;x|s,y} - \beta \I{y;z|s}\\
    \label{eq:CFB}
   &= \I{z;x} - (1 + \beta) \I{y;z|s}.
\end{align}
The Rényi fair IB (RFIB) was introduced as a generalization of IB (for $\zeta=0$) and CFB (for $\rho=0$) in~\citep{Gronowski_RFIB}, where the authors proposed minimizing the following Lagrangian:
\begin{equation}
    \mathcal{L}_\text{RFIB} := \I{z;x} - \rho \I{z;y} - \zeta \I{z;y|s}
\end{equation}
While the last two terms were approximated using cross-entropy, the authors proposed loosening the well-known variational bound $\I{z;x}\le\E_\p{x,y}\D(\p{z|x} \| \q{z})$ (cf.~Section~\ref{sec:CPFSI:variational}) using Rényi divergence and differential Rényi cross-entropy, respectively.

Learning invariant representations $\z$ without a concrete downstream task usually requires that $\z$ contains information about $\x$ (i.e., is a faithful or high fidelity representation) while being independent of a sensitive attribute $\s$. In the context of statistical data privacy, this allows us to share a maximally useful representation $\z$ without leaking information about $\s$. The Privacy-Funnel (PF)~\citep{makhdoumi2014information} addresses the desiderata above. More precisely, the PF Lagrangian formulation aims for $\p{z|x}$ that maximizes $\mathcal L_\text{PF} := (1-\delta)\I{z;x} - \delta\I{s;z}$, where $\delta$ trades between privacy and utility.
Alternatively,~\citet{rodriguez2021variational} introduced the Conditional Privacy Funnel (CPF) with a corresponding neural-based approximation for the related objective. Namely, the CPF Lagrangian minimizes
\begin{subequations}\label{eq:CPF}
    \begin{align}
    \mathcal L_\text{CPF} :&= \I{s;z} - \gamma \I{z; x | s}\\ 
    &= \I{z;x} - (1 + \gamma) \I{z;x|s}.
\end{align}
\end{subequations}

Much like IBSI extends IB, our work extends CPF with side-information (hence CPFSI) and thus generalizes the CPF to learning representations with supervision. Thus, the CPFSI tries to learn invariant but faithful representations that are useful for a downstream task, such as the semi-supervised version of the variational fair autoencoder~\citep{louizos2015variational} or the works of~\citet{zemel2013learning} and~\citet{edwards2016censoring}.

Finally, our work has some similarities to the Flexibly Fair Variational Autoencoder~\citep{creager2019flexibly} and the family of methods included in the Mutual Information-based Fair Representations framework~\citep{song2019learning}. These methods also propose learning fair representation in terms of promoting statistical parity by minimizing a mutual information term. 
Unlike the later examples, our approach is based on variational bounds that obviate the need for adversarial training an information-theoretic objective.

\section{Conditional Privacy-Funnel with Side-Information}
\label{sec:CPFSI}

\begin{figure}[!ht]
    \centering
    \includegraphics[width=.33\textwidth]{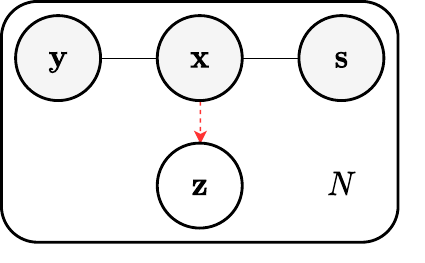}
    \caption{The graphical model that captures the modeling and Markov relations of the CPFSI method and also of the IBSI~\citep{chechik2002extracting}. The solid lines represent the generative assumptions, while the dotted red arrow represents the encoding relation between the input $\x$ and the representation $\z$.}
    \label{fig:pgm}
\end{figure}

CPFSI combines three conflicting goals: Invariance, fidelity, and utility for a downstream task.
Specifically, we would like to learn a representation $\z$ of the covariates $\x$, with the constraint that $\z$ contains as little information as possible about the sensitive attribute $\s$, i.e., we aim for $\z$ being independent of $\s$ or $\z \bot \s$. 
While a faithful representation $\z$ innately contains much information about $\x$, we additionally ensure good performance on a selected downstream task by rewarding representations that contain sufficient information about $\y$.
The covariates $\x$, target $\y$, and sensitive attribute $\s$ are observable quantities, while $\z$ is obtained by encoding $\x$. Thus, the training data contains realizations of $\x$, $\y$, and $\s$, which we use to learn our model.

\subsection{Lagrangian Formulation}
We capture the three aims of CPFSI via information-theoretic quantities. Specifically, we measure the invariance of the representation $\z$ w.r.t.\ the sensitive attribute $\s$ via $\I{s;z}$, its fidelity regarding the covariates $\x$ via $\I{x;z|s}$, and its utility for the downstream task via $\I{y;z|s}$, respectively, where we condition on $\s$ to avoid conflict with the aim for invariance. With this, the Lagrangian formulation of the multi-objective optimization problem can be stated as follows:

\begin{subequations}
\begin{definition}[CPFSI]\label{def:CPFSI}
The Conditional Privacy-Funnel with Side-Information (CPFSI) Lagrangian problem is 
\begin{equation}\label{eq:lagrangian}
\min_{\p{z|x}} \set{
\mathcal L_\mathrm{CPFSI} := \I{s;z} - \gamma \I{z;x|s} - \beta \I{y;z|s}},
\end{equation}	
with $\gamma, \beta \geq 0$.
\end{definition}

Note that~\eqref{eq:lagrangian} differs from~\eqref{eq:CPF} by considering side-information via the term $\I{y;z|s}$. We thus recover the CPF of~\citet{rodriguez2021variational} by setting $\beta=0$. Using the Markov relations (Figure~\ref{fig:pgm}) and the chain rule of mutual information, it is easy to show that 
$\I{s;z} = \I{z;x} - \I{z;x|s}$. 
Hence, we can rewrite the above objective as
\begin{align}\label{eq:lagrangian_adapted}
 	\mathcal L_\text{CPFSI} 
 	= \I{z;x} - \alpha \I{z;x|s} - \beta \I{y;z|s},
 \end{align}
with $\alpha=\gamma+1\ge 1$ and $\beta\ge 0$. 
\end{subequations}

To avoid confusion, we note that a \emph{funnel} or \emph{information bottleneck} in the context of this and other works does not need to be an architectural bottleneck. Indeed, the encoder is not a deterministic function, but a stochastic map, and hence it limits the mutual information between the input and latent space. The “width” of this bottleneck is determined by the Lagrangian parameters during optimization, and not by the dimension of the latent space. Hence, it is valid to call a latent space an information bottleneck even if its dimensionality is larger than the dimensionality of the dataset.

\subsection{Variational Bounds}
\label{sec:CPFSI:variational}

Similar to~\citet{alemi2017deep}, we can upper bound the mutual information between $\x$ and $\z$ using the non-negativity of the Kullback-Leibler (KL) divergence
\begin{align*}
\I{z;x} &:= \E_\p{x,z} \log \frac{\p{z|x}}{\p{z}}\\
        &\leq \E_\p{x,z} \log \frac{\p{z|x}}{\q{z}} \\
        &=    \E_\p{x,y}\D(\p{z|x} \| \q{z}),
\end{align*}
where $\q{z}$ is a variational approximation to the marginal distribution $\p{z}=\E_\p{x} \p{z|x}$.

We are looking for lower bounding the remaining mutual information terms. Still using the non-negativity of the KL divergence, we have
\begin{align*} 
    \I{z;x|s}  &:= \E_\p{x,s,z} \log \frac{\p{x|z,s}}{\p{x|s}} \\
               &\geq \E_\p{x,s,z} \log \frac{\q{x|z,s}}{\p{x|s}}\\
               &= \E_{\p{x,s}\p{z|x}} \left[ \log \q{x|z,s} \right] + \H{x|s},
\end{align*}
where $\q{x|z,s}$ is a variational approximation to the posterior $\p{x|z,s}$. Likewise, we have $\I{y;z|s} \geq \E_\p{s, y, z} \left[ \log\q{y|z,s} \right] + \H{y|s}$.
We can adapt this model for scalable learning and inference analogously to the VAE~\citep{kingma2014autoencoding,rezende2014stochastic}.
First, we drop from the optimization problem the terms that do not depend on the latent variable $\z$, such as $\H{x|s}$ and $\H{y|s}$. Second, we parameterize this stochastic model with deep neural networks with parameters $\theta$ and $\phi$ that will be optimized jointly. Specifically, the parameters $\theta$ belong to a stochastic encoder $\pp{\vec{z|x}}$, while the parameters $\phi$ belong to the variational approximations to the posteriors: The decoder $\qq{\vec{x|z,s}}$ and the predictive posterior $\qq{\vec{y|z,s}}$. The use of amortized inference of $\z$ allows exploiting the similarity between inputs to efficiently encode the underlying observations. 

In addition, we apply the reparameterization trick~\citep{kingma2014autoencoding,rezende2014stochastic} to $\z$, effectively allowing gradient backpropagation through the stochastic layers. 
The reparameterization trick consists of assuming that $\z$ can be decomposed into a stochastic part that does not depend on the model parameters, and a deterministic part that depends on the input $\x$. Assuming $\z \sim \mathcal N(\mu, \sigma \mathbf I)$, we would have $z = \mu +  \sigma \odot n$ with $\n\sim \mathcal N(\mathbf 0, \mathbf I)$.
Finally, we assume that $\p{x,y,s}$ coincides with the empirical distribution  of the dataset $\mathcal D$, which contains realizations $x$, $y$, and $s$, and we approximate the expectations over the representation $\z$ with Monte Carlo sampling, effectively replacing all the above expectations with averages.
In practice, we train our models with mini-batch gradient descent, adopt a deep latent Gaussian model with $\q{z}$ following a standard multivariate normal with zero mean and identity covariance, and use a single sample $z$ from the posterior for training. 

Based on these choices, we substitute the Lagrangian in~\eqref{eq:lagrangian_adapted} by the variational bound estimated with Monte Carlo sampling, obtaining
\begin{multline}\label{eq:variational}
\tilde {\mathcal L}_\text{CPFSI} 
:= \\
\frac 1{|\mathcal D|} \sum_{(x,y,s)\in \mathcal D} \sum_{z\sim \pp{\mathbf z|x}}
\D(\pp{\mathbf z|x} \| q(\mathbf z)) -  \alpha \log \qq{x|z,s} - \beta \log\qq{y|z,s},
\end{multline}
where $|\mathcal D|$ is the dataset size. The variational bound~\eqref{eq:variational} is minimized over the network parameters $\theta$ and $\phi$ of the encoder and variational posteriors, respectively.

The choice of decoder distributions was tailored to the data type of each reconstructed feature: Bernoulli distribution for the binary features; Categorical distribution for nominal features; and Gaussian with a fixed variance parameter for the numerical features. The resulting loss function is the equal weighting combination.


We note that our approximation to the objective conditions on the sensitive attribute $\s$ in the predictive posterior, unlike IBSI's variational approximation in~\citep[Section~2.2]{moyer2018invariant}. This means that our model emphasizes the distinction between $\z$ and $\s$ in the predictor and has the additional feature of allowing interventions in the representation space, allowing for \textit{counterfactual predictions} (cf. Section~\ref{sec:results:classifier}).

\subsection{Conditional Predictive Posterior}
\label{sec:conditional}

Conditioning on the sensitive attribute $\s$ to predict the target $\y$ sets apart our CPFSI, CPF, and CFB~\citep{rodriguez2021variational} from IBSI~\citep{moyer2018invariant}.
On the one hand, providing information from $\s$ eliminates the necessity for $\z$ to learn superfluous data about $\s$ from the input, and should benefit the reconstruction of $\x$ and the classification of $\y$ on the decoder and predictive posterior, respectively. In other words, this conditioning on $\s$ enables an invariant representation $\z$ w.r.t.\  $\s$. Whether conditioning this way suffices to ensure that a representation $\z$ is invariant w.r.t.\ $\s$ is debatable~\citep{tishby2015deep,saxe2018on}. One would have to show that neural network training is inherently compressing, thus eliminating unnecessary information from $\z$ (such as information about $\s$, that is provided directly).
On the other hand, a representation $\z$, obtained by providing $\s$ to a decoder and predictive posterior, may fail to attain satisfactory reconstruction or classification accuracy if the actual decoder and classifier are not provided with the sensitive attribute at inference time. Thus, apparently, the balance between fairness/invariance and downstream task performance should be influenced by the conditioning of variational distributions on the sensitive attribute.

\newcommand{\Enc}{\mathcal{E}}
\newcommand{\Clas}{\mathcal{C}}
To make this intuition precise, we investigate theoretically the case where we are only interested in classification, i.e., a setting similar to the CFB. For the sake of argument, we endow the neural network implementing the classifier sufficient expressive power. Here, the hypothesis class $\Clas$ contains the true posteriors, i.e., $\p{y|z,s},\,\p{y|z}\in\Clas$.
With this, we compare the following two optimization problems over the set $\Enc$ of encoders:
\begin{subequations}
\begin{equation}\label{eq:opt1}
    \min_{p(\z|\x)\in\Enc}\quad I(\z;\s)\quad \text{s.t. } H(\y|\z,\s) \le \varepsilon
\end{equation}
\begin{equation}\label{eq:opt2}
    \min_{p(\z|\x)\in\Enc}\quad I(\z;\s)\quad \text{s.t. }H(\y|\z)\le \varepsilon.
\end{equation}
\end{subequations}
This constraint is similar to stopping training as soon as the classification error of the predictive posterior on the validation set falls below a threshold.

Since $H(\y|\z,\s)\le H(\y|\z)$ (conditioning reduces entropy), the feasible set for the former optimization problem in~\eqref{eq:opt1} is a superset of the feasible set of~\eqref{eq:opt2}. Hence, conditioning the predictive posterior on the sensitive attribute $\s$ achieves at least as small $I(\s;\z)$ as not conditioning. At the same time, $H(\y|\z,\s)\le\varepsilon$ \emph{does not} imply that $H(\y|\z)\le\varepsilon$. Thus, the error probability of a classifier that relies on $\z$ alone and that is not provided with the sensitive attribute $\s$ may be larger for $\z$ obtained from solving~\eqref{eq:opt1} than for $\z$ obtained from solving~\eqref{eq:opt2}.

The last statement needs to be qualified: There is no one-to-one correspondence between conditional entropy and error probability, there are only bounds (e.g.,~\citep{Han_Fano,Feder_Fano}). However, we believe that aside from the theoretical argument provided via conditional entropy, the last statement is also plausible from a learning perspective: Optimizing~\eqref{eq:opt2} ensures that $\z$ is useful for classification \emph{on its own}, at least for a classifier with the same architecture as the predictive posterior $\q{y|z}$. In contrast, the $\z$ obtained by solving~\eqref{eq:opt1} is \emph{not necessarily} useful on its own, not even for a classifier with the same expressive power as the variational classifier $\q{y|z,s}$. Information about $\y$ contained in $\z$ may be accessible only with side-information $\s$.

\subsection{Semi-supervised Learning} 
In this study, we tailored the objective to learn with a few labels with a simple strategy: We set $\beta=0$ for observations without corresponding target label. This additional unsupervised loss term is then summed to the supervised loss, scaled by $\max(|\mathcal B_u|/|\mathcal B_s|, 1)$, where $|\mathcal B_u|$ and $|\mathcal B_s|$ represent the unsupervised and supervised batch sizes, respectively. 
A similar strategy can be applied to the variational approximation of IBSI in~\citep{moyer2018invariant}.

\section{Funnels and Bottlenecks }\label{sec:funck}

In the previous section we introduced CPFSI~\eqref{eq:lagrangian_adapted}, which requires $\alpha=\gamma+1\ge 1$ and $\beta\ge 0$.
It is simple to relax these conditions and require only that $\alpha,\beta\ge 0$. We note that, given an appropriate choice of parameters, the CPFSI, CPF, and CFB are all equivalent to the following FUNnel and bottleneCK (FUNCK) objective:
\begin{equation}\label{eq:delta_lagrangian}
\mathcal L_\text{FUNCK} :=
(1-\delta)\I{z;x} +  \delta \I{s;z} - \gamma \I{z;x|s} - \beta \I{y;z|s}.
\end{equation}
Setting $\delta=1$, we recover~\eqref{eq:lagrangian}, i.e., CPFSI. With $\delta=0$ and $\beta\ge 1$ we recover~\eqref{eq:CFB} from CFB, while for $\delta=0$ and $\gamma\ge 1$ (or for $\delta=1$ and $\gamma\ge 0$) we recover~\eqref{eq:CPF} from the CPF.

Apparently, for $\delta<1$, the first term of~\eqref{eq:delta_lagrangian} antagonizes the reconstruction term $\I{z;x|s}$, especially if $\gamma$ is small. This allows us to connect FUNCK with IBSI (and CLUB), which aims for compressed representations rather than for representations that allow high-fidelty reconstruction. Recalling the IBSI objective:
\begin{align*}
    \mathcal L_\text{IBSI}&= \I{z;x} + \gamma \I{s;z} - \lambda \I{y;z}\\
    &= \I{z;x} - \alpha \I{z;x|s} - \beta \I{y;z},
\end{align*}
we see that $\mathcal L_\text{IBSI}$ can be recovered from $\mathcal L_\text{FUNCK}$ by setting $\delta=0$ and $\gamma\in [0,1)$, or by setting $\delta\in(0,1)$ and $\gamma=0$. The main difference between FUNCK and IBSI/CLUB in this case remains that FUNCK conditions on $\s$ for evaluating the utility of $\z$ w.r.t.\ classifying $\y$ (i.e., $\I{y;z|s}$), while IBSI and CLUB do not (i.e., $\I{y;z}$). The effects of this difference deserve separate attention and are briefly discussed in Section~\ref{sec:conditional}.

\section{Experimental Setup} 
\label{sec:methodology}


\subsection{Datasets}
We performed experiments on four benchmark datasets extensively used in the fields of algorithmic fairness and private representation learning. Each of these datasets --- Adult~\citep{becker1996adult}, Dutch~\citep{laan2000census}, Credit~\citep{yeh2016creditcard}, and COMPAS~\citep{propublica2016compas} --- is tabular, featuring a binary target variable $\y$ and binary sensitive attribute $\s$. 

The target task in the Adult dataset is predicting if a person's income is larger than \$50k. This dataset is one of the most commonly used dataset in algorithmic fairness work and has around 48k observations and 14 features. 
The Dutch dataset target is a binary attribute related with the person's occupation. This dataset has 60420 observations and 10 features.
Both these datasets consist of census data and gender is the protected or sensitive attribute.
The Credit dataset has 30k observations, 23 features, and the task is to predict if the client will miss the payment in the following month. Gender is the sensitive attribute.
Finally, in the ProPublica COMPAS dataset,
the target is recidivism within two years. The sensitive attribute is ethnicity, and the dataset has only approximately 6k observations and 12 features.

As part of the dataset preprocessing, we applied one-hot encoding to categorical features and standardized numerical features across all datasets. Since the Dutch dataset did not contain numerical features, we created one based on target encoding of the age attribute in the training set.


\subsection{Models} 
We performed experiments on the variational approximations of IBSI~\citep{moyer2018invariant} and different instances of the FUNCK family of objectives described in Sections~\ref{sec:relatedwork} and~\ref{sec:CPFSI}. These are the CPF and CFB~\citep{rodriguez2021variational}, and CFPSI (Ours).
For semi-supervised learning, we performed experiments on a semi-supervised version of CPFSI and the Variational Fair Autoencoder (VFAE,~\citep{louizos2015variational}).
All models were implemented with similar architectural choices (which are detailed in the accompanying code) and latent representation with 32 dimensions. Both the encoder and decoder are feed-forward neural networks with Rectified Linear Units (ReLU) activation functions, while the classifier is a logistic layer from the representation to the target attribute.  Further details are provided along with the supplementary code. Regarding IBSI, we did not follow the exact implementation from~\citet[Section~2.1.1]{moyer2018invariant}. Instead, to enable a fair comparison between the objectives (and to avoid being confounded by effects due to their KL divergence estimator), we implemented IBSI using similar approximations as our other methods, 
essentially following~\citet{alemi2017deep}. Most hyperparameters were informed by the related work.

In the SSL setting, the level of supervision ranged from 4 to 256 labeled examples per class in semi-supervised experiments. The reconstruction term weight was set to 1 for all semi-supervised experiments, while the weighting for the target cross-entropy term varied. The VFAE was implemented without the Maximum-Mean Discrepancy (MMD) term, which enforces invariance. 

\subsection{Training and Evaluation}

The training objectives for all models are influenced by hyperparameters $\alpha$ and $\beta$ that control the reconstruction and classification losses, respectively. 
All losses were weighted by two terms ($\alpha$ and $\beta$) that controlled the reconstruction and classification losses, respectively. The KL term was implicitly controlled by the total magnitude of both terms.
These hyperparameters ranged from 1 to 1024 and were generated from the sequence $2^{2k}, k=0,\ldots,5$. The CFB and CPF models had $\alpha$ and $\beta$ to zero, respectively. For IBSI~\eqref{eq:IBSI}, $\alpha$ ranged between 0 and 1 --- 0.001, 0.01, 0.1, 0.5 and 1.

All methods were trained for a maximum of 200 epochs using the Adam optimizer, learning rate of $0.001$, and batches of (at most) size 256. We reduced the learning rate when the validation loss demonstrated no improvements for at most 10 epochs. After 20 epochs without improvements, we stopped training. 

Each experiment was repeated five times using 5-fold cross-validation for distinct training and evaluation splits. The validation set was obtained from the training set resulting in an 18:2:5 (72:8:20) partition ratio for training, validation, and testing, respectively. 

\subsubsection{Metrics}

The binary classifiers $f(\cdot)$ were evaluated in terms of accuracy, and discrimination~\citep{zemel2013learning}. The reconstruction error of the numerical features was measured in terms of the mean-absolute error (MAE).
Specifically, for a dataset $\mathcal D=\set{(x_i,y_i,s_i)}_{i=1}^N$
the $i$-th observation has binary label $y_i$, binary sensitive attributes $s_i$, and representation $z_i$ of input $x_i$; let $N_C$ be the number of observations in the dataset when restricted to the condition $C$, e.g. $s=1$.
\begin{definition}[Discrimination]
The discrimination or statistical parity gap,
\begin{equation*}
    \Delta_\text{disc} := \left|\frac{\sum_{i:\,s_i=0} f(z_i)}{ N_{s=0}} - \frac{\sum_{i:\,s_i=1} f(z_i)}{ N_{s=1}} \right|,
\end{equation*}
measures independence between a binary classifier $f(\cdot)$ of $y_i$ and a sensitive attribute $s_i$.
\end{definition}

\renewcommand{\S}{\mathcal S}
\renewcommand{\F}{\mathcal F_\mathrm{ref}}
The following multi-objective performance indicators were used to compare the method's performance across multiple conflicting objectives. Namely, they measure the quality of Pareto front approximations produced by the evaluation of the representations produced by the studied methods. Let $\S$ be a Pareto front approximation and $\F$ be a reference Pareto front~\citep{hansen1994evaluating, audet2021performance}:
\begin{definition}[C1R]
The ratio of reference points (C1R) metric quantifies the proportion of reference points that are covered by the approximation of the Pareto front. The C1R metric is give by 
\begin{align*}
\mathrm{C1R}(\S;\F) := \frac{|\{p \in \S: p \in \F\}|}{|\F|}.
\end{align*}
\end{definition}
C1R allows us to assess the degree to which our method's solutions align with a predefined set of desirable outcomes, emphasizing the utility aspect of the trade-off.

\begin{definition}[C2R]
The ratio of non-dominated points by the reference set (C2R) metric calculates the proportion of solutions that are not dominated by the reference set:
\begin{align*}
\mathrm{C2R}(\S;\F) := \frac{|\{p \in \S: \nexists\; r \in \F\;\mathrm{s.t.\ }\; r \leq p\}|}{|\S|},
\end{align*}
where $\leq$ means that $r$ dominates $p$. In other words, $r$ is not worse than $p$ in all objectives and better in at least one objective.\footnote{The definition of dominance via the symbol $\leq$ is derived from multi-objective minimization problems.}
\end{definition}
C2R offers an insight into the invariance of our method's performance, revealing how well it maintains quality across different scenarios.

\begin{definition}[HV,~\citep{zitzler1999evolutionary}]
The hypervolume indicator, also named S-metric, is described as the volume of the space in the objective space dominated by the Pareto front approximation $\S$ and delimited from above by a reference objective vector $r \in \F$ such that for all $p \in \S, p\leq r$. The hypervolume indicator is given by 
\begin{align*}
    \mathrm{HV}(\S; r) :=\lambda_m\left(\bigcup_{p\in \S} [p, r]\right),
\end{align*}
where $\lambda_m$ is the $m$-dimensional Lebesgue measure. 
\end{definition}
In the context of our work, the hypervolume serves as a quantitative measure of how well our method's solutions dominate the trade-off space.

For each method studied, the solution set is the Cartesian product of all representations' evaluations obtained from the different estimators. The reference Pareto front is the combination of all these evaluations for all different seeds.

\subsection{Evaluating Representations}
\label{sec:methodology:representation}

Algorithms \ref{alg:repr-privacy-fairness} and \ref{alg:repr-reconstruction} describes broadly the evaluation procedure. Using a similar approach to~\citep{louizos2015variational} with learners with varying degrees of flexibility, we trained Logistic Regression (LR) and Random Forest (RF) learners to predict either the target or sensitive attribute from the generated representations $z$. These were probed in terms of their classification error, different notions of invariance (a.k.a.\ ``fairness" or ``privacy"), and reconstruction error (related to faithfulness or fidelity to the input). We refer to these learners as \textit{estimators}. This was done to measure, linearly and non-linearly, the information content left by the encoders. In other words, these estimators were used as measurement instruments for the fairness and privacy of $z$. To control for overfitting on the test set, we repeated the evaluation five times on different splits of the test set and reported in the figures the median of these measurements.
The demographic parity of these (binary) classifiers was examined through estimates of discrimination~\citep{zemel2013learning}. 
The evaluation of the representation's fidelity to the input was restricted to a single numerical feature. All RF predictors used 100 estimators, and the RF regressors were limited to a maximum depth of 8.
The learners' evaluation was repeated five times using a stratified cross-validation scheme on the test set representations. All evaluated models were picked using the best validation loss, and all reported results are obtained from the test set's representations. The trade-off between classification error, invariance, and reconstruction fidelity --- achieved by sweeping $\alpha,\beta$ --- was summarized using multi-objective performance indicators~\citep{audet2021performance}: 
the ratio of points a method achieves in a reference Pareto front (C1R); the ratio of points non-dominated by the reference (C2R); and the hypervolume of the set in objective space that is dominated by the method.

Since all measurements were obtained from different partitions of each test dataset which were aggregated afterward, the results' variability for each method is explained by the different weight's initialization, different training set splits, and the different configuration for the objectives in terms of $\alpha$ and $\beta$. Baseline values were calculated by adapting the evaluation above to the original dataset \textit{without} $s$ as a covariate.


\begin{algorithm}[htb]
\caption{Representation Evaluation -- Fairness and Privacy.}\label{alg:alg1}
\begin{algorithmic}[1]
\State \textbf{Input:}  $\textsc{Encoder}_\theta, s,y$ \Comment{Encoder, sensitive attribute, 
target}
\State \textbf{Output:} evaluation
\State  evaluation $\gets \textsc{List}()$
\State $z \gets \textsc{Encoder}_\theta(x)$ \Comment{Representation dataset}
\State \textbf{for} $(z_\text{tr}, y_\text{tr}, s_\text{tr}), (z_\text{te}, y_\text{te}, s_\text{te})$ \textbf{in} $\textsc{KFold}(z, y,s, k=5)$ \textbf{do}
\State \hspace{1em} \textbf{for} $t_\text{tr}$ \textbf{in} $[y_\text{tr}, s_\text{tr}]$ \textbf{do}
\State \hspace{2em} \textbf{for} \textsc{Estimator} \textbf{in} $[\textsc{LogisticRegression}, \textsc{RandomForest}]$ \textbf{do}
\State \hspace{3em} \textsc{Estimator}.fit$(z_\text{tr}, t_\text{tr})$
\State \hspace{3em} evaluation $\texttt{extend}$ \textsc{Evaluator}$(\textsc{Estimator}(z_\text{te}), y_\text{te}[, s_\text{te}])$
\end{algorithmic}
\label{alg:repr-privacy-fairness}
\end{algorithm}

\begin{algorithm}[htb]
\caption{Representation Evaluation -- Fidelity.}\label{alg:alg2}
\begin{algorithmic}[1]
\State \textbf{Input:} $\textsc{Encoder}_\theta, x, j, s$ \Comment{Encoder, dataset, feature index, sensitive attribute}
\State \textbf{Output:} evaluation
\State evaluation $\gets \textsc{List}()$
\State $z \gets \textsc{Encoder}_\theta(x)$
\State \textbf{for} $(z_\text{tr}, x_{j,\text{tr}}), (z_\text{te}, x_{j,\text{te}})$ \textbf{in} $\textsc{KFold}(z, x_j, k=5)$ \textbf{do}
\State \hspace{1em} \textbf{for} \textsc{Estimator} \textbf{in} $[\textsc{LogisticRegression}, \textsc{RandomForest}]$ \textbf{do}
\State \hspace{2em} \textsc{Estimator}.fit$(z_\text{tr}, x_{j,\text{tr}})$
\State \hspace{2em} evaluation $\texttt{extend}$ \textsc{Evaluator}$(\textsc{Estimator}(z_\text{te}), x_{j,\text{te}})$
\end{algorithmic}
\label{alg:repr-reconstruction}
\end{algorithm}

\subsection{Fair Classification using the Predictive Posterior}

In addition to assessing the fairness of the latent representation, the experiments also evaluate the learned neural predictive posterior of both the fully and semi-supervised models. The focus is on a form of fair classification where intervention occurs at test time in the observed $\s$. This is utilized to examine the impact of counterfactual predictions on discrimination. However, the objective is not to make causal inferences or to predict the outcomes of actual interventions, but to yield fairer predictions on observational data.
We explored two straightforward interventions on $s$: flipping the binary value of $s$, and fixing $s$ to $0.5$. This approach is described in Algorithm~\ref{alg:posterior}.

\begin{algorithm}[!htpb]
\caption{Predictive Posterior Evaluation.}
\begin{algorithmic}[1]
\State \textbf{Input:} 
$\textsc{Encoder}_\theta$, $x$, $s$ \Comment{Encoder, original dataset, sensitive attribute}
\State \textbf{Output:} evaluation
\State evaluation $\gets \textsc{List}()$
\State $z \gets \textsc{Encoder}_\theta(x)$
\State \textbf{for} $s'$ \textbf{in} $[s, 1-s, 1/2]$ \textbf{do}
\State \hspace{1em} evaluation \texttt{extend} $\textsc{Evaluator}(\textsc{Posterior}_\phi, z[, s'])$
\end{algorithmic}
\label{alg:posterior}
\end{algorithm}
\subsection{Computing Infrastructure}
The experiments were conducted using the following computing infrastructure:
40 CPUs, 256 GB RAM, NVIDIA Tesla P100 16 GB, on Ubuntu 20.04 LTS. All code was written for Python 3.8. Remaining software and more details of our experiments are provided along with the supplementary code.

\section{Results}
\label{sec:results}

In this section, we highlight our key findings in the fully and semi-supervised settings for invariant representation learning and fair classification. Supplementary results can be found in 
Appendix~\ref{sec:appendix:results}.

\subsection{Representation Invariance}
\label{sec:results:invariance} 

\begin{figure*}[tb]
    \centering
    \includegraphics[width=.9\linewidth]{"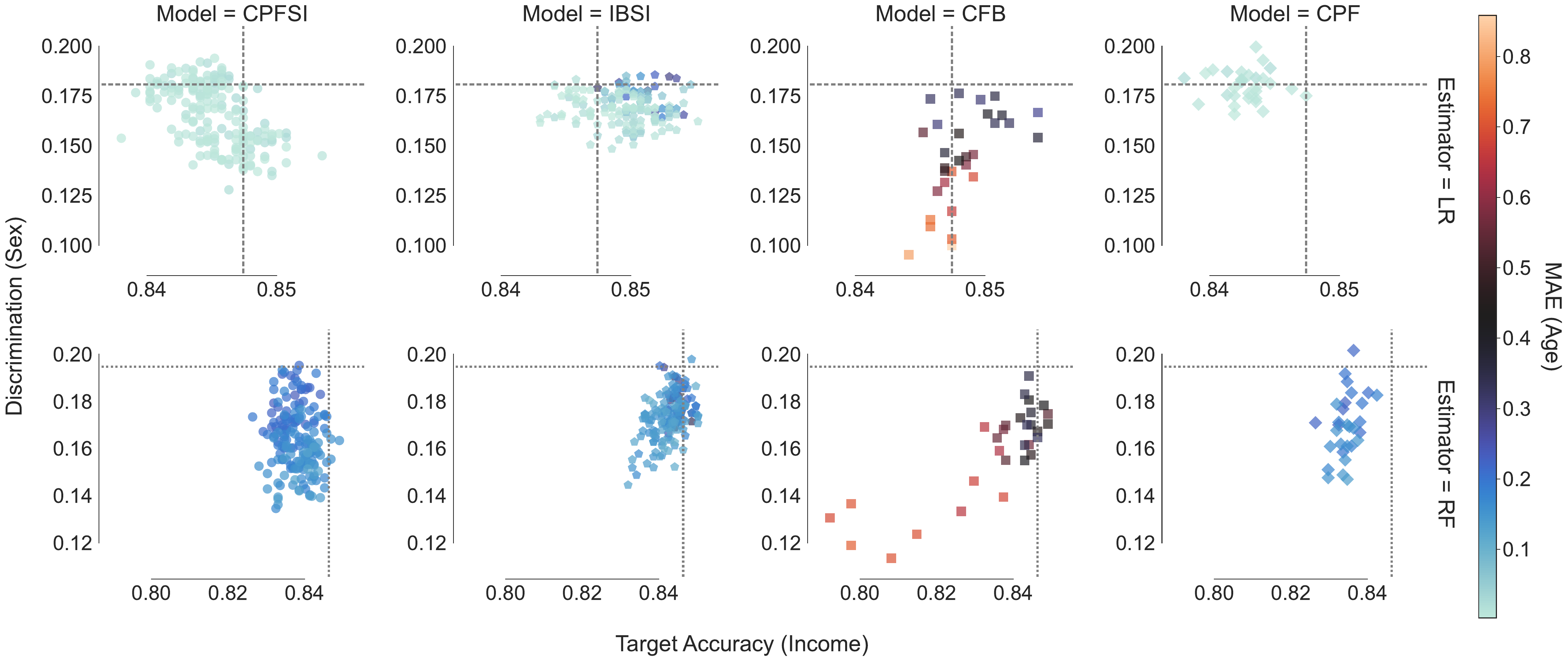"}
    \caption{Representation's fairness 
    on the \textit{Adult} dataset. The dashed line marks the estimators --- Logistic Regression (top) and Random Forest (bottom) --- evaluated on the original features. The color scale represents the mean absolute reconstruction error from the Linear Regression and Random Forest Regressor from $z$ of a numerical feature of $x$, across configurations obtained by sweeping $\alpha$ and $\beta$, illustrating the three-way evaluation of utility, invariance, and representation fidelity.}
    \label{fig:representation-fairness:adult}
\end{figure*}
\begin{figure}[!htbp]
    \centering
    \includegraphics[
        width=\linewidth]{"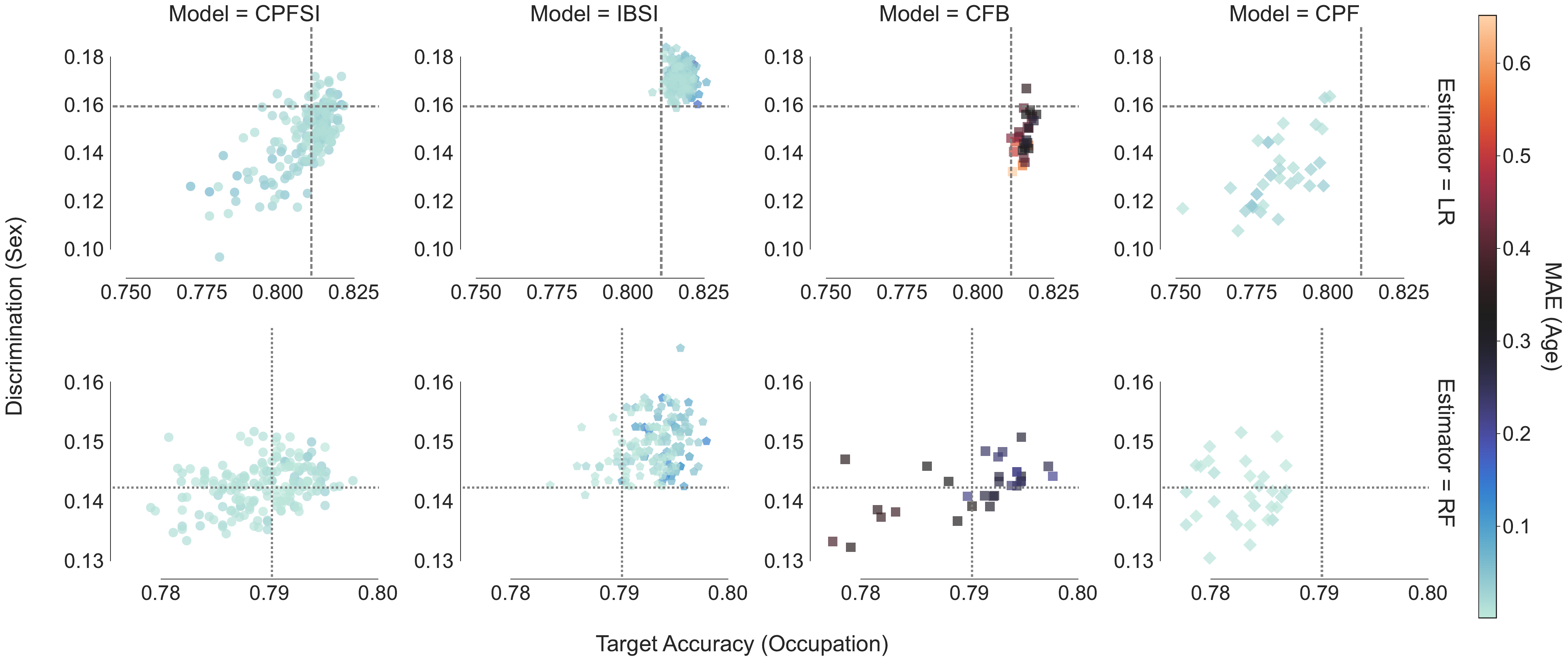"}
        \caption{Representation's fairness 
        on the \textit{Dutch} dataset. 
        } 
        \label{fig:appendix:representation-fairness-dutch}
\end{figure}

Figure~\ref{fig:representation-fairness:adult} characterizes the behavior of the different methods on the fully-supervised case and shows to what extent the different variational approaches control the utility-fairness-fidelity trade-off of the representations generated from the test set of the Adult dataset.
Most methods have configurations that matched or exceeded the baselines accuracies with lower discrimination, which demonstrates the applicability of these techniques for data debiasing. The lowest discrimination is achieved by CFB; since this method learns exclusively task-specific representations, this comes at the cost of fidelity of the reconstruction. 
Representations obtained via IBSI are most predictive of the target attribute, but suffer from higher discrimination than CPFSI.
This can be verified both in the linear (LR) and nonlinear (RF) evaluations of the representations. 
This observation is most obvious in the Dutch dataset (Figure~\ref{fig:appendix:representation-fairness-dutch}). In contrast, CPFSI has both high accuracy and low discrimination while encoding the input data with high fidelity. Hence, the CPFSI learns task-agnostic representations that can be useful in application settings where there are potentially other downstream tasks.
\begin{figure}[!htbp]
    \centering
    \includegraphics[
        width=\linewidth]{"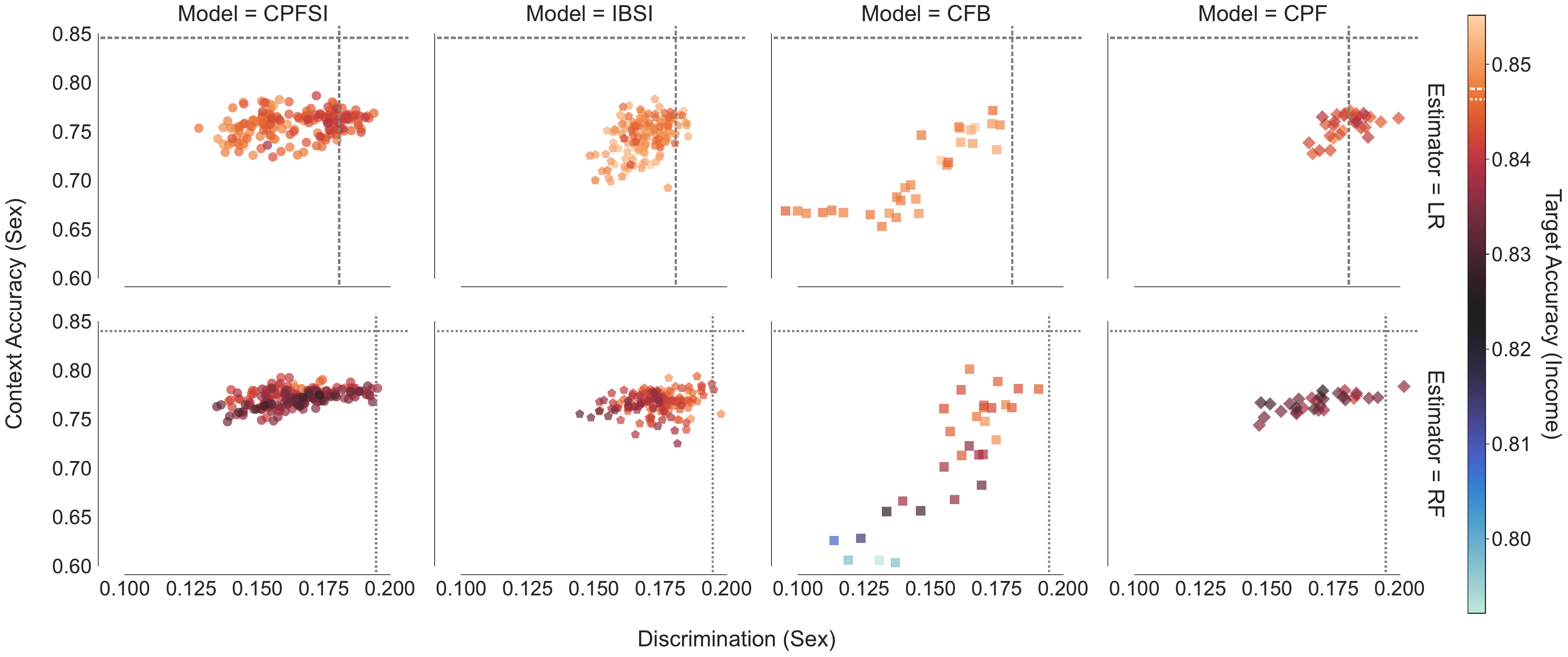"}
        \caption{Representation's fairness vs.\ privacy 
        on the \textit{Adult} dataset.
        } 
        \label{fig:appendix:representation-fairness-privacy-adult}
\end{figure}
However, Figure~\ref{fig:appendix:representation-fairness-privacy-adult} shows that both CPF and CPFSI have little control on the utility-privacy trade-off, while CFB has the best control of the utility-privacy trade-off despite targeting fairness. This result questions the applicability of these privacy-funnel methods for their intended purpose in private representation learning.

Table~\ref{tab:multiobjective-metrics} summarizes results for all datasets as a function of the representation quality in terms of reconstruction fidelity, fairness/privacy, and target accuracy in comparison to a reference set approximated from all methods and random seeds evaluations. Our results indicate that CPFSI and IBSI achieve competitive performance with the state-of-the-art methods and generally achieve good utility-invariance-fidelity trade-offs.
For the fair representations, CPFSI consistently presents the largest percentage of points on, or non-dominated by, the reference set (C1R and C2R, respectively). The same holds for IBSI in terms of privacy-preserving representations. The comparably poor performance of CFB and CPF is probably caused by the fact that these objectives do not consider reconstruction and target accuracy, respectively. The hypervolume characterizes how much of the trade-off is explored by the methods. The comparably large standard error can be attributed to the different weight initialization, but also reflects that the experiment replicates were tested on disjoint splits of the same dataset.
Generally, the results were clearer for larger datasets, while it appears harder to differentiate between methods for small datasets, as in~\citep{moyer2018invariant}.
The results for Adult, Dutch, Credit, and COMPAS datasets are summarized in terms of the three objectives in Table~\ref{tab:multiobjective-metrics}. 

\renewcommand*{\b}[1]{\mathbf{#1}}
\begin{table}[!t]
    \centering
    \resizebox{.99\columnwidth}{!}{
    \begin{tabular}{@{}cccrrr@{}}
    \toprule
    Dataset &
    Property & 
    Method &
    C1R (\%) &
    C2R (\%) &
    HV ($\times 100$) \\ \midrule
    \multirow{8}{*}{Adult} &
    \multirow{4}{*}{Fairness} &
    CPF   & 
    $0.00\pm 0.00$ &
    $0.00\pm 0.00$ &
    $70.43\pm 0.49$ \\
    & &
    CFB &  
    $4.86\pm 0.90$ &
    $30.67\pm 5.28$ &
    $51.64\pm 0.65$ \\
    & &
    IBSI & 
    $\b{7.57}\pm 2.58$ &
    $18.79\pm 5.60$ &
    $72.28\pm 0.16$ \\
    & &
    CPFSI & 
    $\b{7.57}\pm 2.34$ &
    $\b{38.73}\pm 12.88$ &
    $\b{72.98}\pm 0.33$ \\ \cmidrule(l){2-6} 
    &
    \multirow{4}{*}{Privacy} &
    CPF   &  
    $0.48\pm 0.43$ &
    $6.67\pm 5.96$ &
    $21.65\pm 0.49$ \\
    & &
    CFB & 
    $5.71\pm 0.85$ &
    $\b{40.00}\pm 5.96$ &
    $19.66\pm 0.38$ \\
    & & 
    IBSI & 
    $\b{9.05}\pm 1.70$ &
    $33.96\pm 9.27$ &
    $\b{25.33}\pm 0.22$ \\
    & &
    CPFSI & 
    $4.76\pm 2.78$ &
    $20.82\pm 11.73$ &
    $22.01\pm 0.46$ \\
    \midrule
    \multirow{8}{*}{Dutch} &
    \multirow{4}{*}{Fairness} &
    CPF   &  
    $2.26\pm 0.63$ &
    $20.38\pm 6.46$ &
    $69.74\pm 0.25$ \\
    & & 
    CFB &  
    $1.89\pm 0.53$ &
    $18.10\pm 5.29$ &
    $55.46\pm 0.23$ \\
    & & 
    IBSI & 
    $6.04\pm 2.09$ &
    $12.69\pm 4.11$ &
    $70.33\pm 0.06$ \\
    & & 
    CPFSI & 
    $\b{9.81}\pm 2.64$ &
    $\b{31.75}\pm 8.65$ &
    $\b{72.14}\pm 0.42$ \\ \cmidrule(l){2-6} 
    &
    \multirow{4}{*}{Privacy}
    &
    CPF   &  
    $0.98\pm 0.87$ &
    $12.00\pm 10.73$ &
    $30.47\pm 0.32$ \\
    & &
    CFB & 
    $4.88\pm 0.69$ &
    $36.00\pm 4.94$ &
    $26.89\pm 0.08$ \\
    & &
    IBSI & 
    $\b{10.24}\pm 2.12$ &
    $\b{43.81}\pm 9.66$ &
    $\b{32.86}\pm 0.06$ \\
    & &
    CPFSI & 
    $3.90\pm 1.77$ &
    $26.67\pm 13.62$ &
    $31.86\pm 0.39$ \\
    \midrule
    \multirow{8}{*}{Credit} &
    \multirow{4}{*}{Fairness} &
    CPF   &  
    $0.00\pm 0.00$ &
    $0.00\pm 0.00$ &
    $238.04\pm 0.26$ \\
    & &
    CFB &  
    $0.57\pm 0.51$ &
    $4.00\pm 3.58$ &
    $206.58\pm 1.49$ \\
    & &
    IBSI & 
    $8.00\pm 4.15$ &
    $17.33\pm 8.23$ &
    $\b{241.61}\pm 0.24$ \\
    & &
    CPFSI &  
    $\b{11.43}\pm 3.33$ &
    $\b{31.25}\pm 6.75$ &
    $241.58\pm 0.25$ \\ \cmidrule(l){2-6} 
    &
    \multirow{4}{*}{Privacy} &
    CPF   &  
    $0.43\pm 0.38$ &
    $5.00\pm 4.47$ &
    $101.15\pm 0.36$ \\
    & &
    CFB &  
    $2.13\pm 0.60$ &
    $19.33\pm 5.69$ &
    $97.86\pm 0.63$ \\
    & &
    IBSI &  
    $\b{11.91}\pm 2.14$ &
    $\b{41.86}\pm 6.40$ &
    $\b{109.54}\pm 0.87$ \\
    & &
    CPFSI &  
    $5.53\pm 1.42$ &
    $29.00\pm 7.85$ &
    $103.86\pm 0.49$ \\
    \midrule
    \multirow{8}{*}{COMPAS} &
    \multirow{4}{*}{Fairness} &
    CPF   &  
    $6.15\pm 0.84$ &
    $\b{20.08}\pm 2.73$ &
    $53.89\pm 0.24$ \\
    & &
    CFB &  
    $1.54\pm 0.84$ &
    $4.86\pm 2.73$ &
    $53.76\pm 0.19$ \\
    & &
    IBSI &  
    $4.62\pm 3.34$ &
    $12.54\pm 9.70$ &
    $\b{55.25}\pm 0.42$ \\
    & &
    CPFSI &  
    $\b{7.69}\pm 1.09$ &
    $19.13\pm 4.44$ &
    $55.23\pm 0.10$ \\ \cmidrule(l){2-6} 
    &
    \multirow{4}{*}{Privacy} &
    CPF   & 
    $3.12\pm 0.88$ &
    $24.67\pm 6.06$ &
    $22.60\pm 0.14$ \\
    & &
    CFB & 
    $2.50\pm 0.56$ &
    $18.00\pm 4.77$ &
    $26.26\pm 0.34$ \\
    & &
    IBSI & 
    $\b{10.00}\pm 3.89$ &
    $\b{30.63}\pm 12.94$ &
    $\b{26.35}\pm 0.42$ \\
    & &
    CPFSI & 
    $4.38\pm 1.90$ &
    $25.00\pm 12.47$ &
    $23.28\pm 0.14$ \\
    \bottomrule
    \end{tabular}
    }
    \caption{Characterization of each method's solution set in terms of
    utility, invariance, and reconstruction fidelity, using performance indicators for multi-objective optimization (higher numbers are better).
    Averages and standard errors are computed from different initialization on different splits of the training set.}
    \label{tab:multiobjective-metrics}
\end{table}

\begin{figure*}[!htbp]
    \centering
    \includegraphics[width=\linewidth]{"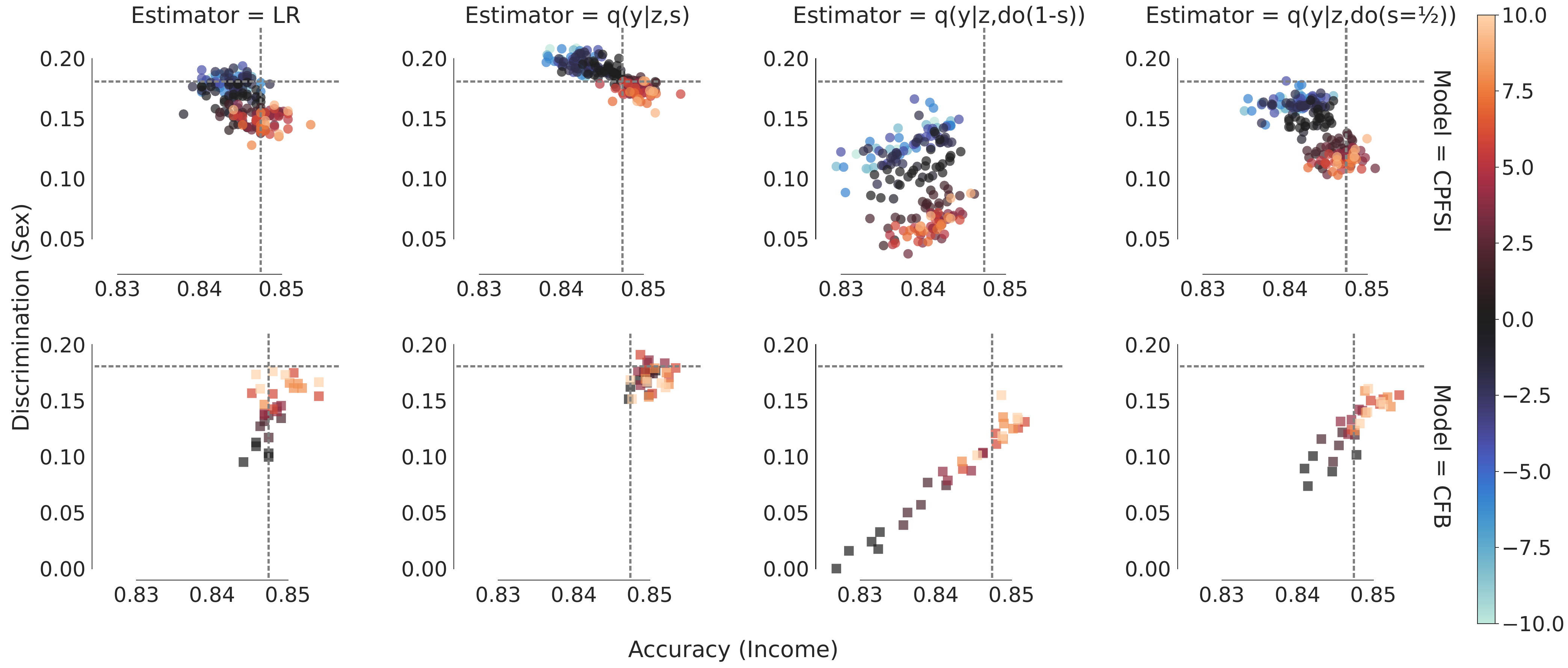"}
    \caption{Comparison between representation fairness and predictive posteriors, for different interventions on the sensitive attribute $\s$, on the \textit{Adult} dataset. The gray-dashed reference line belongs to a Logistic Regression estimator on the original data. Results for other datasets are in Appendix~\ref{sec:appendix:results}.}
    \label{fig:predictor-adult-credit}
\end{figure*}

\subsection{Semi-supervised Learning}

\begin{figure}[!htb]
    \centering
    \includegraphics[width=\linewidth]{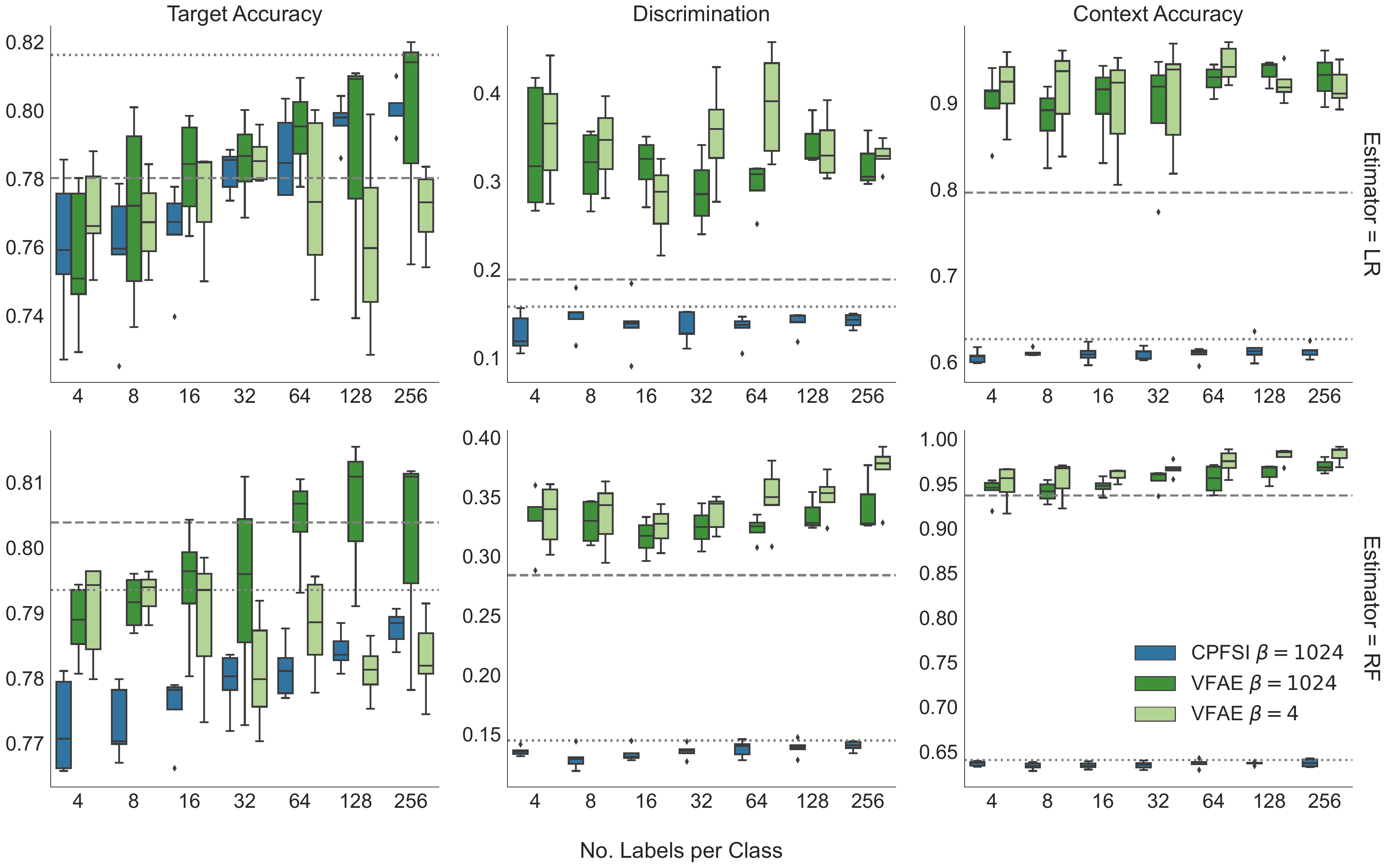}
    \caption{Representation quality for $\alpha=1$ and different numbers of labels on the \textit{Dutch} dataset. The dashed and dotted lines mark the unsupervised (CPF) and fully supervised baselines, respectively. Results for other datasets are contained in Appendix~\ref{sec:appendix:results}.}
    \label{fig:ssl-representation-dutch}
\end{figure}

The SSL results (Figure~\ref{fig:ssl-representation-dutch} and Figures \ref{fig:ssl-representation-adult} to \ref{fig:ssl-representation-compas}) indicate that the CPFSI is capable of improving the learning efficiency for invariant representation learning with just a few labels per class. Compared to the semi-supervised VFAE, CPFSI not only achieves comparable target accuracy, but also reduces discrimination and leakage about the sensitive attribute. For example, on the Dutch dataset (Figure~\ref{fig:ssl-representation-dutch}), a few labels enhance target accuracy while keeping discrimination low. 

The VFAE exhibited substantially higher adversarial accuracy (lower privacy) for the remaining datasets.
Furthermore, CPFSI is simpler to implement and train than the hierarchical (semi-supervised) VFAE. Regarding the remaining datasets, we observe that both discrimination and target accuracy results are comparable, but the information leakage is substantially higher for the VFAE.
\citet[Section 2.2]{louizos2015variational} lead us to conjecture that the VFAE's higher leakages come from encoding $\z_1$ from $\x$ and $\s$ in the variational posterior $\q{z_\text{1}|x,s}$. To mitigate this issue, the authors introduced a MMD term in the objective to encourage an invariant marginal posterior $\p{z_\text{1}|s}$. We did not observe an added benefit in adding a MMD penalty to the CPFSI Lagrangian. Furthermore, we did not encounter a practical benefit on the added complexity of the VFAE's hierarchical model, and the CPFSI is simpler to implement and train.

\subsection{Fair Classification using the Predictive Posterior}
\label{sec:results:classifier}

In this set of experiments, we evaluate the learned predictive posterior $\qq{\vec{y|z,s}}$ for the representation $\z$ and by fixing $\s$ according to two predefined rules (Algorithm~\ref{alg:posterior}), one of which removes the requirement of having the sensitive attribute at test time. 
%
This suggests an alternative approach to fair classification that complements training a separate classifier on the invariant representations.
It consists of finding an invariant representation regarding the protected attributes and a policy to intervene in $\s$ that achieves the most fairness while retaining predictive performance.
This approach is possible for any conditional predictive posteriors and/or classifiers like CPFSI, CFB, and the semi-supervised version of the VFAE.
In Figure~\ref{fig:predictor-adult-credit}, the analysis of the predictive posteriors indicates that one can improve on the representation's fairness by intervening on a conditional classification without loss in accuracy.
Moreover, our results indicate that the best intervention strategy depends on the dataset, but so far, we could not conclusively trace this dependency  to the relation between $\y$ and $\s$. In any case, providing $s=1/2$ the predictive posterior appears to decrease or retain the discrimination while maintaining accuracy, compared to providing the true sensitive label.

\section{Discussion}

In today's data-driven epoch, it remains challenging to create useful data representations that remain invariant to nuisance factors, especially in scarce data and low supervision settings. Information-theoretic objectives are a concise way to conveniently specify multiple requirements from first principles.

In this work, we have expanded on the IBSI, CPF, CFB, and VFAE analyses, contributed with a novel multi-objective analysis never performed on these methods, and proposed the CPFSI, a new method for invariant representation learning in the supervised and semi-supervised setting, which conceptually expands the recently proposed CPF. The introduction of a conditional element to the predictive posterior, gave us another perspective on these methods and on the complex interplay between the different factors involved in invariant representation learning and fair classification.

The concept of representation is generally not well-defined, but it is common to consider that a good representation encapsulates multiple explanatory factors, as noted by~\citet{bengio2013representation}.
High reconstruction fidelity is an indication of the usefulness of the representation in undefined downstream tasks, and this is achieved by CPF, CPFSI, and IBSI. While IBSI originally targets at a single task,~\citet{moyer2018invariant} introduced an artifact in the resulting variational bound that encouraged high-fidelity representations by applying the chain rule for mutual information.
In contrast, CFB, which is also specialized to a single task, has representations with high reconstruction error that make it useless for arbitrary downstream tasks. 
Nevertheless, the simplicity of the CFB allows for better encoding in terms of fairness and privacy. We have identified CPFSI as covering a middle ground between these extremes, while also extending the reachable Pareto front for the conflicting objectives. 

In Figure~\ref{fig:ssl-representation-dutch}, we have shown that our method is capable of learning fairer representations and to reduce leakage with just a few labels. Further, while we focused on the semi-supervised case, CPFSI can easily be extended towards other weakly supervised scenarios or settings where we aim to learn a shared representation that is invariant to some context $\s$. Much like~\citet{louizos2015variational}, we could also extend our work to the domain adaptation setting to study the robustness to shifts of these representation learning methods.

All models were restricted to a particular approximation and architectural choices in the framework of amortized variational inference parameterized by neural networks. As such, considerations about the objectives must be interpreted carefully, since it is not obvious to what extent statements based on the objective can carry over to the variational bound derived from its Lagrangian.

Moreover, our method is also capable of fair classification and achieves lower discrimination than non-conditional models when we intervene on the sensitive attribute. This structural property of our model sets it apart from previous work such as~\citep{moyer2018invariant}. Additionally, the CFB, essentially a conditional version of IBSI, shows potential as a better alternative to IBSI when a high representation fidelity of the input is not a priority.

Finally, despite CPF and CPFSI being designed to control the utility-privacy trade-off, both methods are easier to control in a fairness setting. Besides, our results strongly suggest that the IBSI and the CFB methods have a better control over the representation's privacy.

\section{Limitations} 

Unlike differential privacy~\citep{dwork2006calibrating}, we do not claim that our method provides any strict privacy guarantee for data publishing. We are simply borrowing the terminology from~\citet{makhdoumi2014information,rodriguez2021variational} where privacy loss is understood as the adversary's ability to predict a sensitive attribute from the published representations $\z$. In other words, it reduces the leakage of the published representations $\z$.
In fact, our experiments revealed a substantial leakage by the encoded inputs by the different methods. This is exacerbated by the fact that there is no privacy guarantee for each individual, given that any observation is at risk of revealing some private information.

Selecting the desired trade-off between fairness, privacy, predictive performance, and reconstruction fidelity requires setting $\alpha$ and $\beta$. This non-trivial task is inherent of the multi-objective nature of these problems and depends on the user desires and could the subject of techniques for hyperparameter tuning, like Bayesian optimization.

While we believe that our approximations to the Lagrangian objectives are flexible enough to be accurate, situations where the sensitive attribute $\s$ and the target $\y$ are correlated, could yield random or degenerate representations regarding $\y$ when training an unsupervised model, since removing the influence of $\s$ can also remove all the relevant information in $\z$. Studying the effect of these choices was out-of-the scope of this study.

Industry applications frequently deal with small, tabular datasets~\citep{dua2019uci}. But, Deep Neural Networks (DNNs) generally are most limited when applied to domains characterized by such datasets~\citep{grinsztajn2022why}. Nonetheless, research in fairness, accountability, and transparency has seen an increasing number of work that applies DNNs to these domains. While we demonstrated that CPFSI is competitive with previous work across tabular datasets, 
we believe that even better performance will be possible when combining the CPFSI training objectives with DNN architectures that are specifically designed for tabular data~ \citep{borisov2022deep}.

\section{Conclusion}
The pursuit of trustworthy machine learning remains a key challenge in an era of pervasive data-driven approaches, especially in settings where data is scarce or supervision is minimal. In this work, we build on the VFAE, IBSI, CPF, and CFB frameworks to introduce CPFSI, a new method for invariant representation learning in both supervised and semi-supervised scenarios, which generalizes the recently proposed CPF. Finally, we demonstrate how these methods perform on ordinary tabular datasets under limited data conditions --- situations where their benefits are especially relevant.

\clearpage


\begin{appendices}


\section{More Results}
\label{sec:appendix:results}

\begin{figure}[!htbp]
    \centering
    \includegraphics[
        width=\linewidth]{"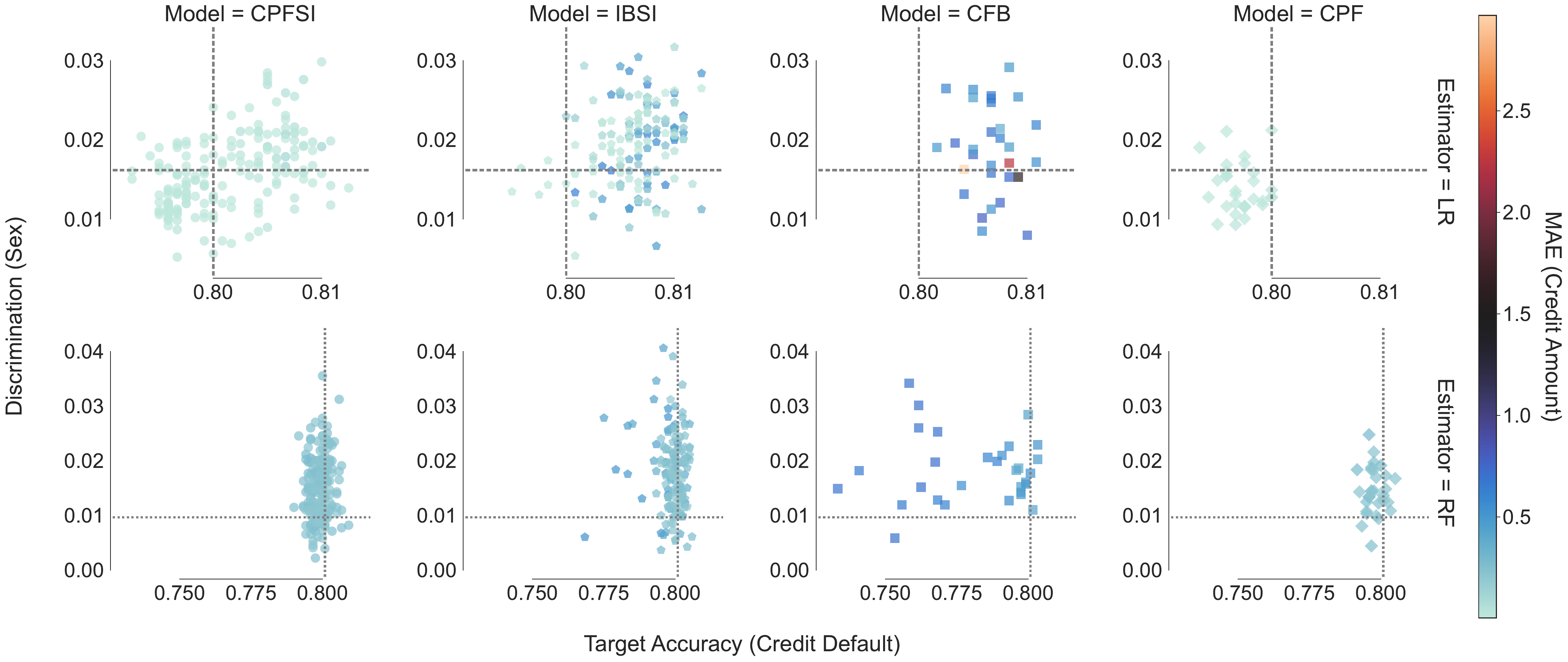"}
        \caption{Representation's fairness on the \textit{Credit} dataset.} 
        \label{fig:appendix:representation-fairness-credit}
\end{figure}

\begin{figure}[!htbp]
    \centering
    \includegraphics[
        width=\linewidth]{"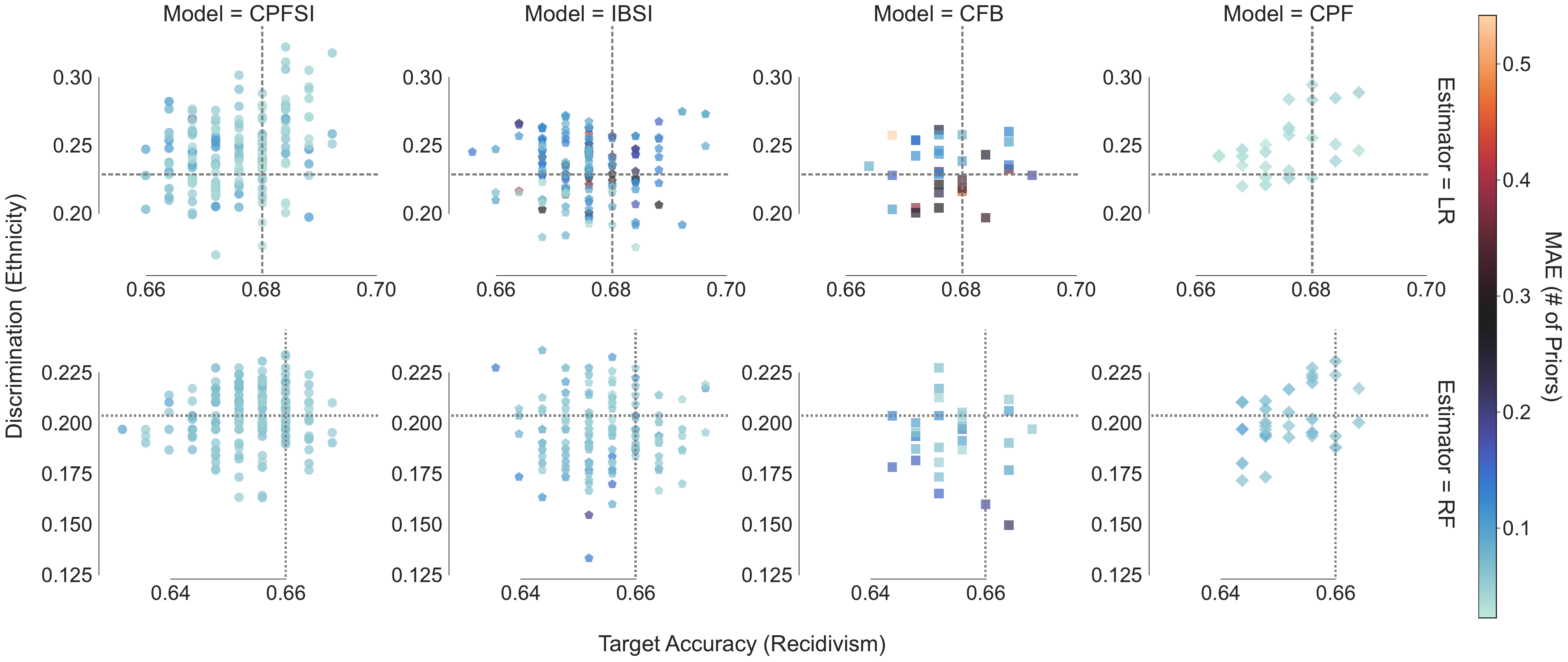"}
        \caption{Representation's fairness on the \textit{COMPAS} dataset.} 
        \label{fig:appendix:representation-fairness-compas}
\end{figure}

\begin{figure}[!htbp]
    \centering
    \includegraphics[width=.9\linewidth]{"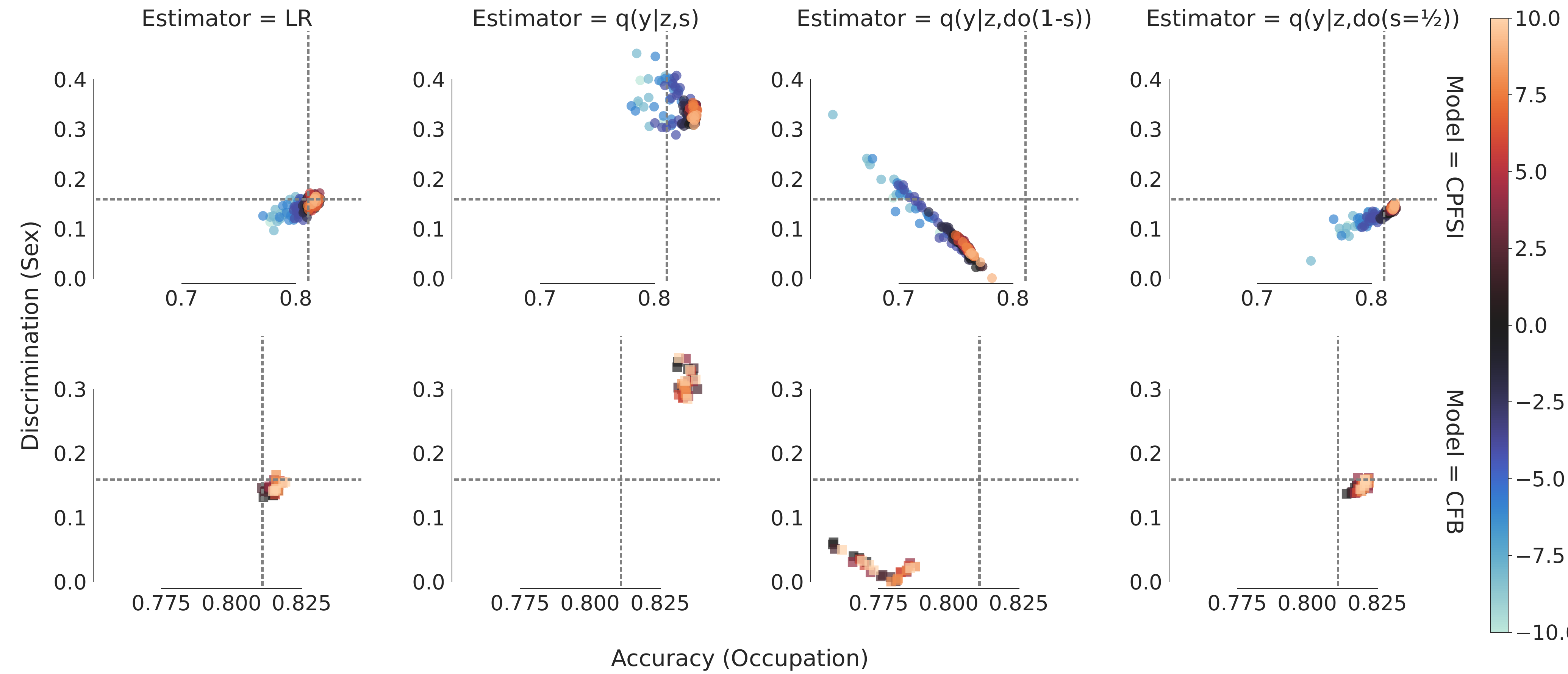"}
    \caption{Comparison between representation fairness and predictive posteriors on the \textit{Dutch} dataset.}
    \label{fig:predictor-dutch}
\end{figure}

\begin{figure}[!htbp]
    \centering
    \includegraphics[width=.9\linewidth]{"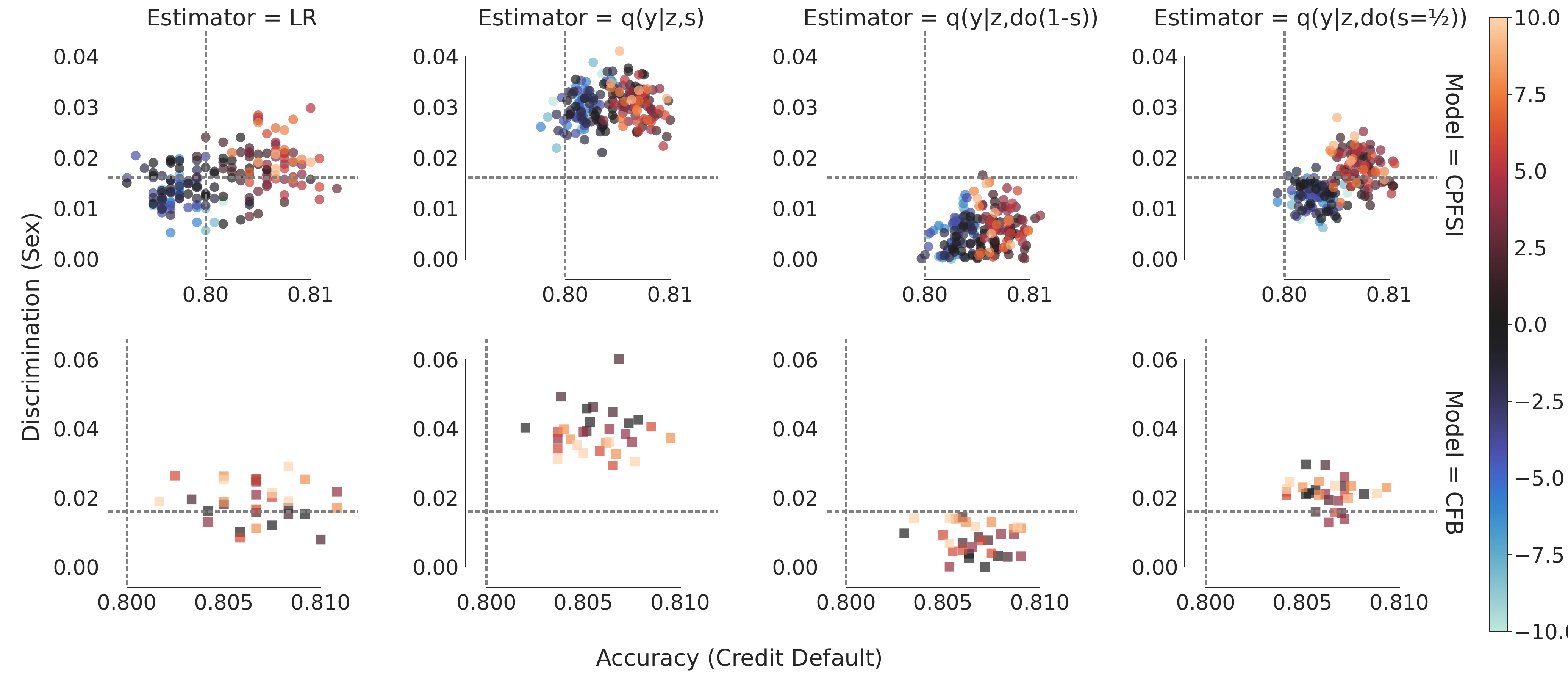"}
    \caption{Comparison between representation fairness and predictive posteriors on the \textit{Credit} dataset.}
    \label{fig:predictor-credit}
\end{figure}

\begin{figure}[!htbp]
    \centering
    \includegraphics[width=.9\linewidth]{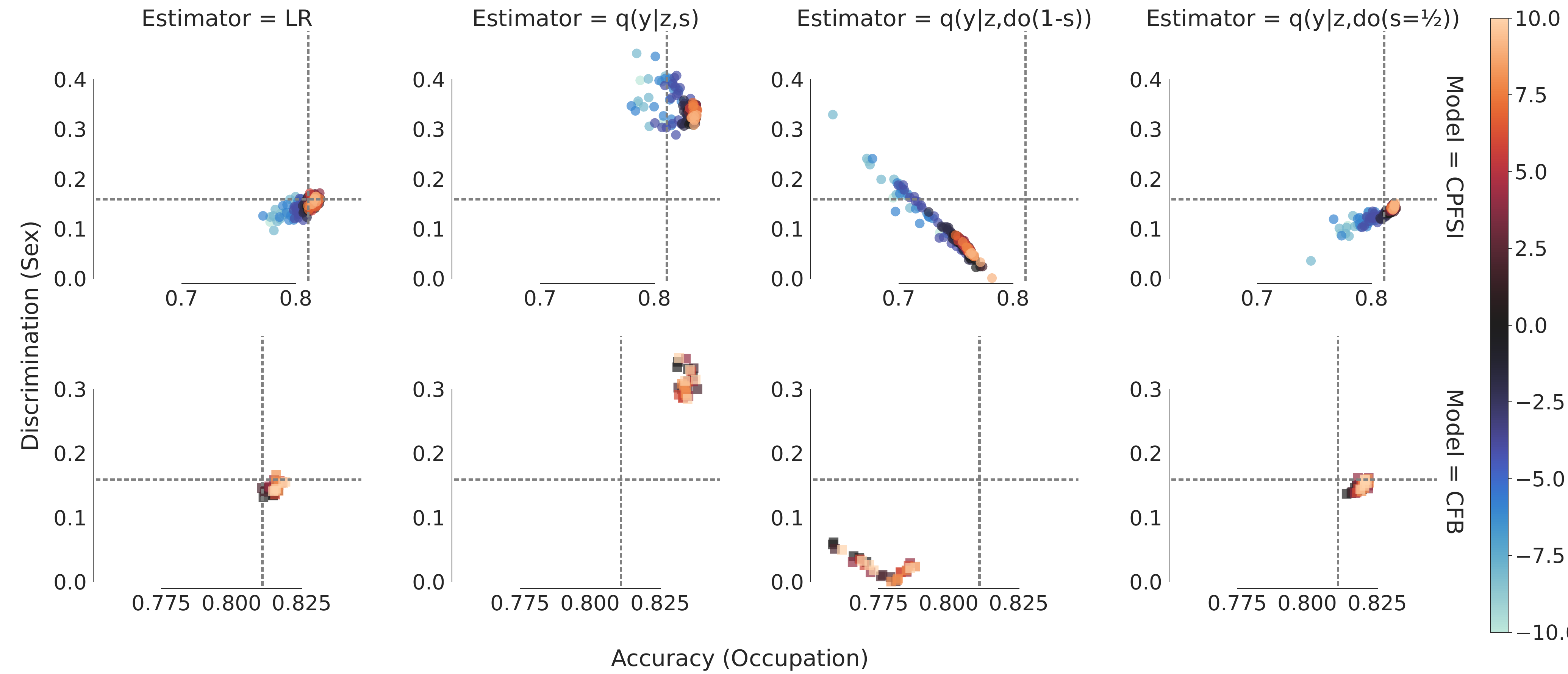}
    \caption{Comparison between representation fairness and predictive posteriors on the \textit{COMPAS} dataset.}
    \label{fig:predictor-compas}
\end{figure}

\begin{figure}[!htbp]
    \centering
    \includegraphics[
        width=\linewidth]{"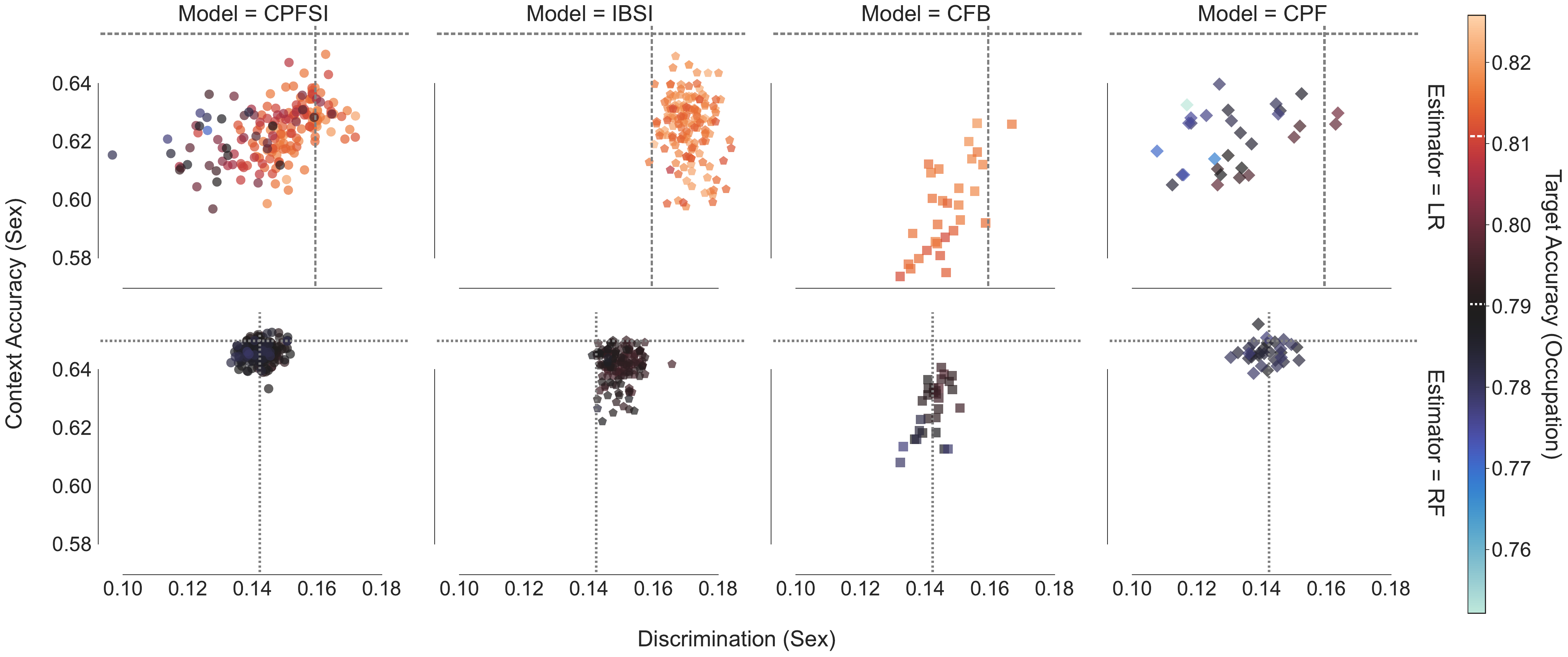"}
        \caption{Representation's fairness vs.\ privacy on the \textit{Dutch} dataset.} 
        \label{fig:appendix:representation-fairness-privacy-dutch}
\end{figure}

\begin{figure}[!htbp]
    \centering
    \includegraphics[
        width=\linewidth]{"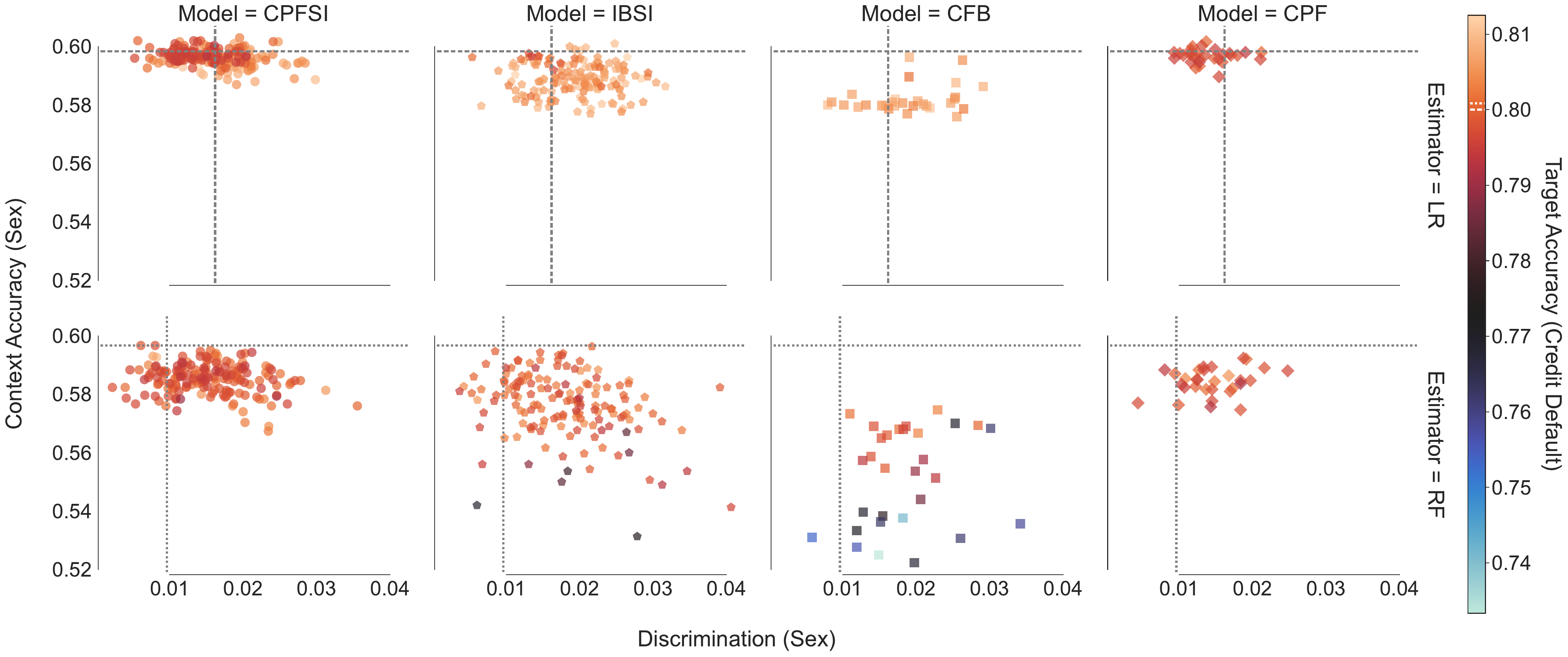"}
        \caption{Representation's fairness vs.\ privacy on the \textit{Credit} dataset.} 
        \label{fig:appendix:representation-fairness-privacy-credit}
\end{figure}

\begin{figure}[!htbp]
    \centering
    \includegraphics[
        width=\linewidth]{"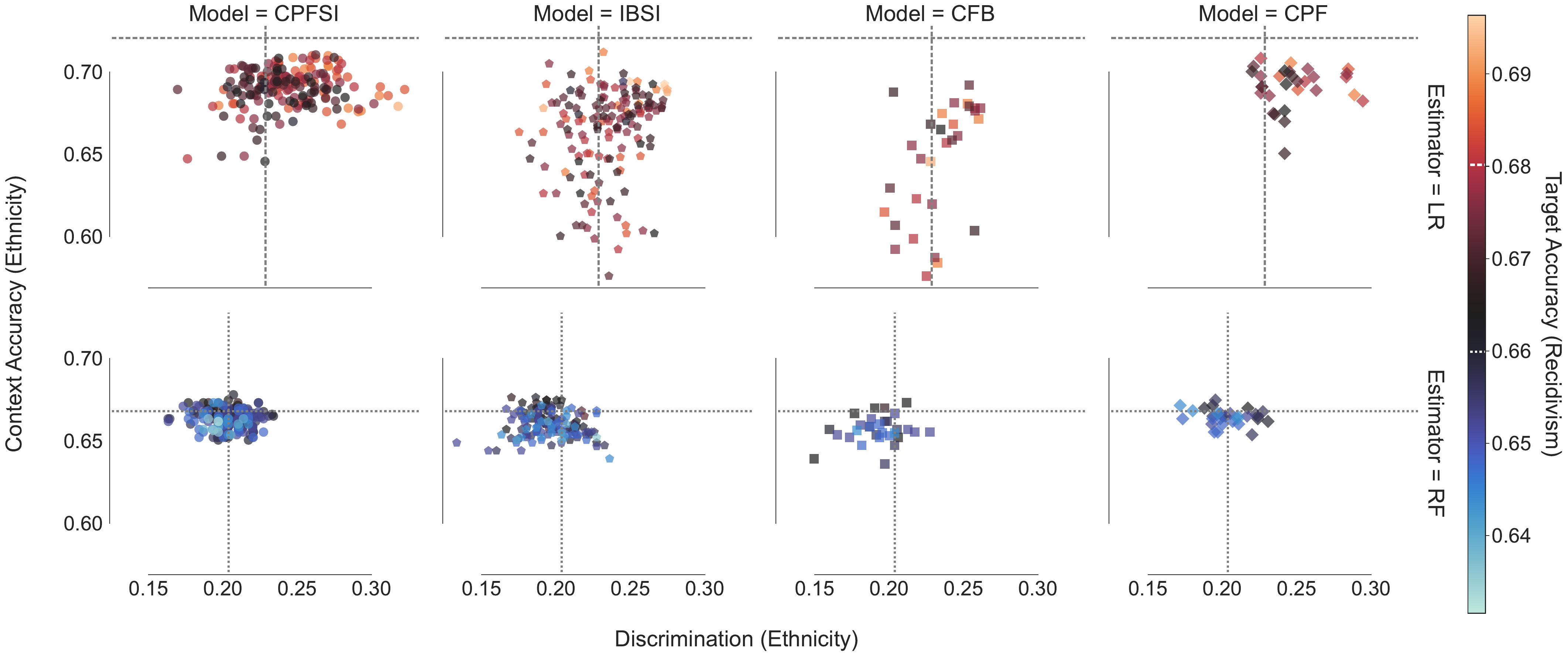"}
        \caption{Representation's fairness vs.\ privacy on the \textit{COMPAS} dataset.} 
        \label{fig:appendix:representation-fairness-privacy-compas}
\end{figure}

\begin{figure}[!htb]
    \centering
    \includegraphics[width=.85\linewidth]{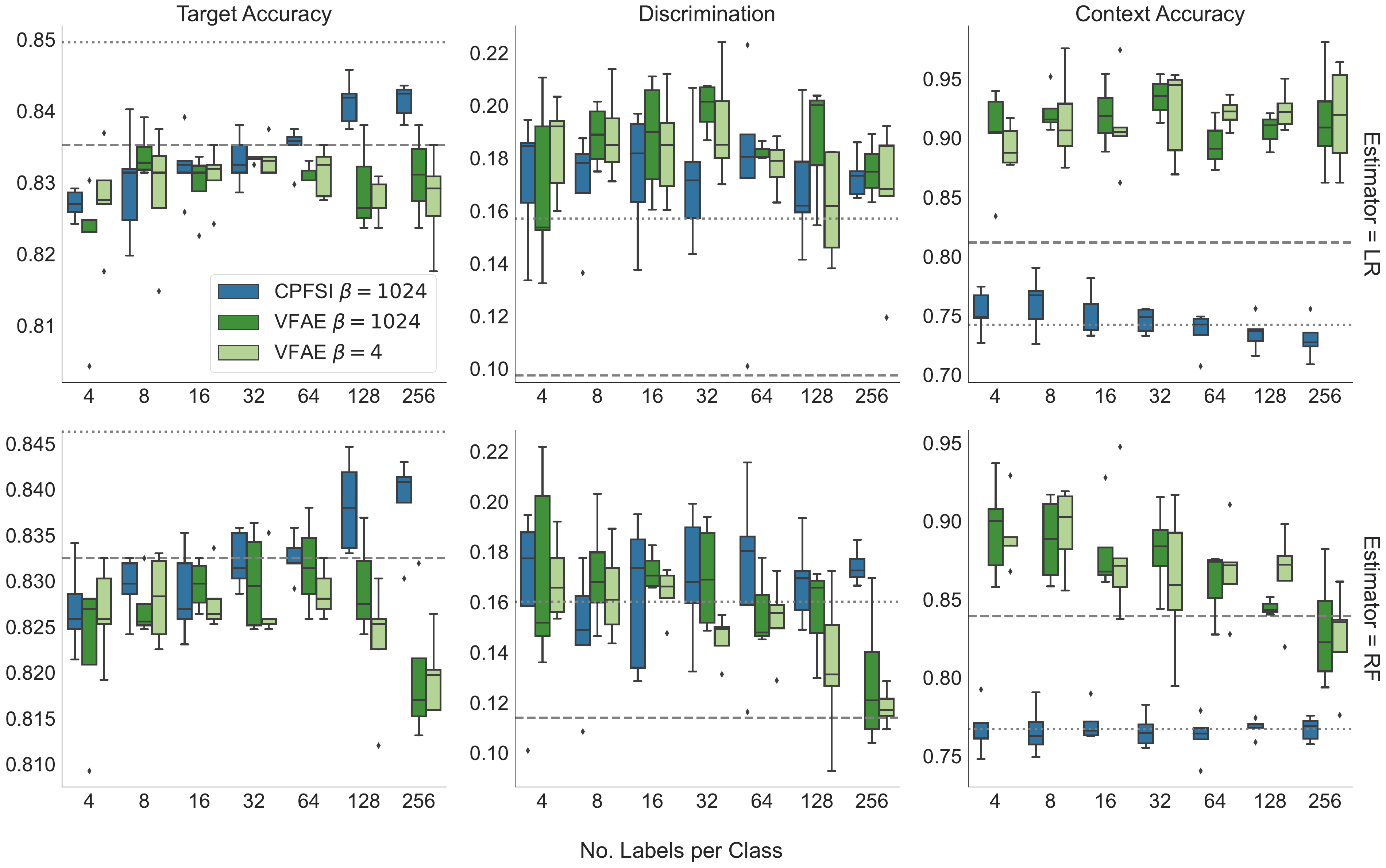}
    \caption{Representation quality for the SSL on the \textit{Adult} dataset. }
    \label{fig:ssl-representation-adult}
\end{figure}

\begin{figure}[!htb]
    \centering
    \includegraphics[width=.85\linewidth]{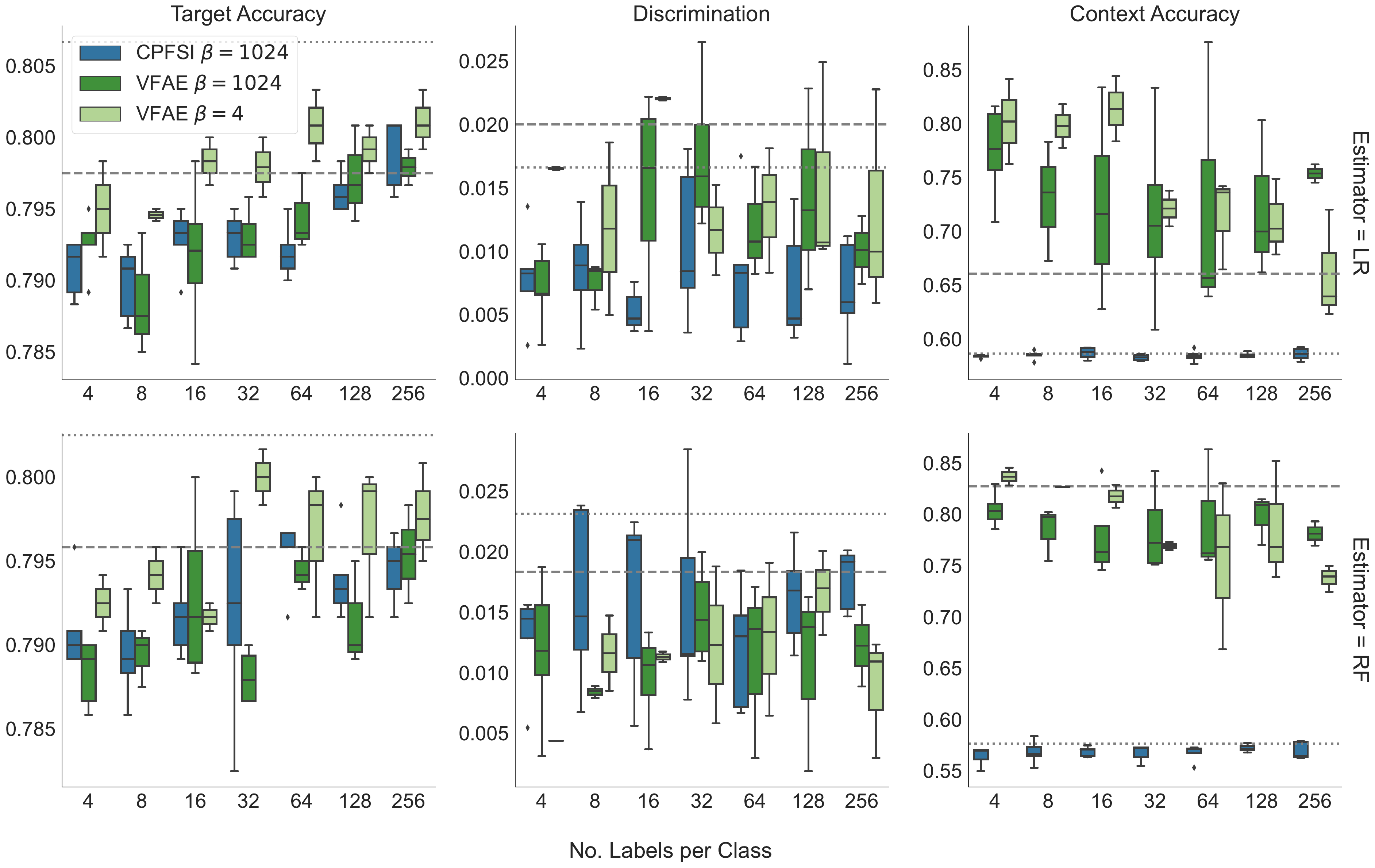}
    \caption{Representation quality for the SSL on the \textit{Credit} dataset.}
    \label{fig:ssl-representation-credit}
\end{figure}

\begin{figure}[!htb]
    \centering
    \includegraphics[width=.85\linewidth]{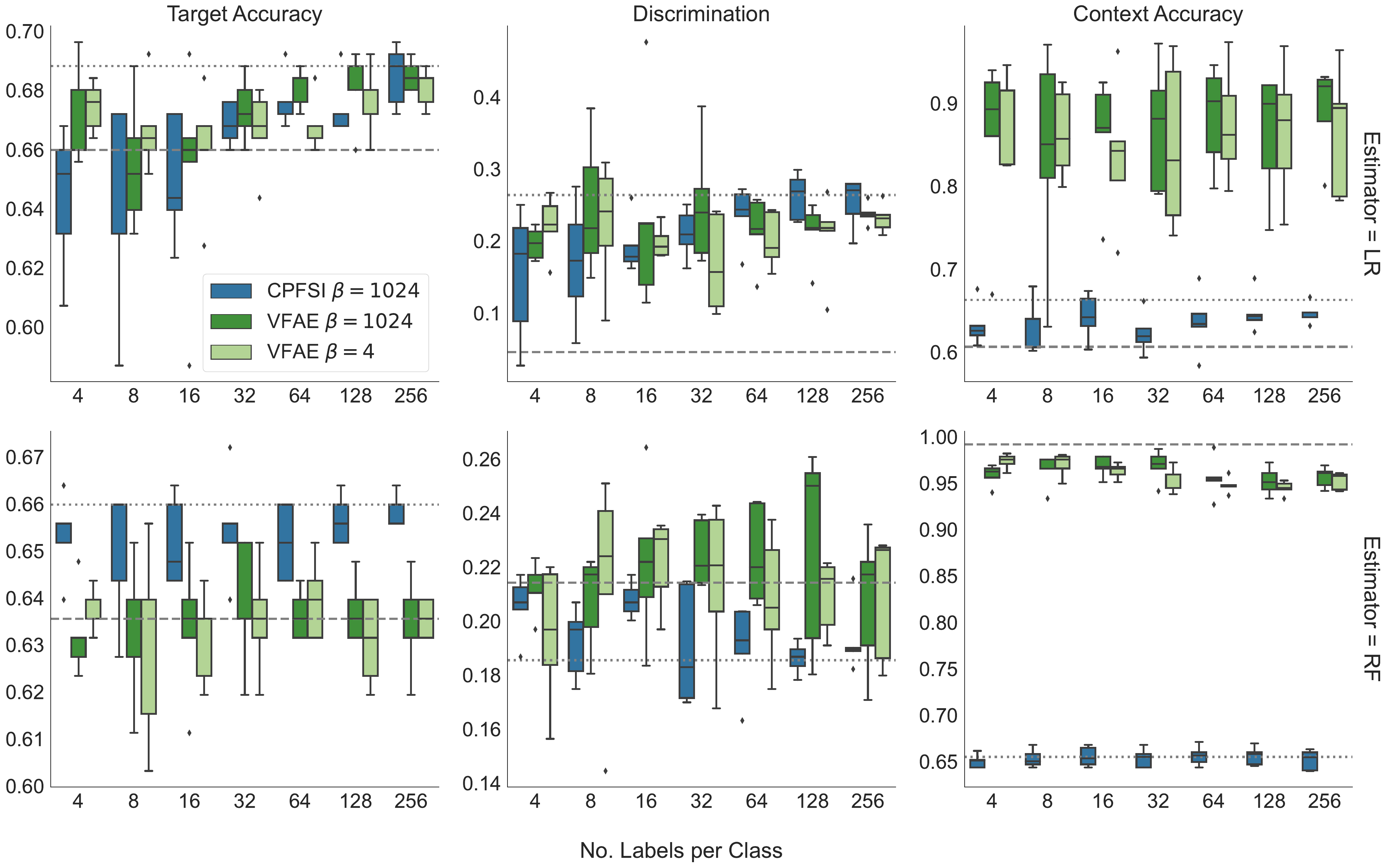}
    \caption{Representation quality for the SSL on the \textit{COMPAS} dataset.}
    \label{fig:ssl-representation-compas}
\end{figure}

In Figures~\ref{fig:appendix:representation-fairness-credit} and~\ref{fig:appendix:representation-fairness-compas}, 
the color scale represents reconstruction error from $z$ a numerical feature $x$, using a linear and random forest regressors (top and bottom rows). The dashed lines correspond to the baseline of the majority classifier, while the dotted lines correspond to the baseline of the row estimator --- Logistic Regression (LR, top) and Random Forest (RF, bottom) --- on the original features.
%
Figures~\ref{fig:predictor-dutch} to~\ref{fig:predictor-compas} compare interventions on the sensitive attribute $\s$. The gray-dashed reference line belongs to a LR estimator on the original data
The color scale represents the magnitude of the hyperparameters $\alpha$ and $\beta$ that control the representation's compression.
%
Figures~\ref{fig:appendix:representation-fairness-privacy-dutch} to~\ref{fig:appendix:representation-fairness-privacy-compas} compare the fairness and privacy of the representations. 
%
% ...
%
Figures~\ref{fig:ssl-representation-adult} to~\ref{fig:ssl-representation-compas} summarize the results in SSL setting for $\alpha=1$ and different numbers of labels. The dashed and dotted lines mark the unsupervised (CPF) and fully supervised baselines, respectively.






\end{appendices}

\section*{Declarations}

\bmhead{Funding}
This work was supported by the ``DDAI" COMET Module within the COMET --- Competence Centers for Excellent Technologies Programme, funded by the Austrian Federal Ministry (BMK and BMDW), the Austrian Research Promotion Agency (FFG), the province of Styria (SFG) and partners from industry and academia. The COMET Programme is managed by FFG. The financial support by the Austrian Federal Ministry of Labour and Economy, the National Foundation for Research, Technology and
Development and the Christian Doppler Research Association is gratefully
acknowledged. Furthermore, the research was funded by RHI Magnesita.

\bmhead{Conflicts of interest/Competing interests} None.

\bmhead{Ethics approval} Not Applicable.

\bmhead{Consent to participate} Not Applicable.

\bmhead{Consent for publication} All authors who participated in this paper give the publisher the permission to publish this work.

\bmhead{Availability of data and material} All data and software is open-source.

\bmhead{Code availability} Code implementation is publicly available at: \href{https://github.com/jmachadofreitas/funck-ml2024}{github.com/jmachadofreitas/funck-ml2024}

\bmhead{Authors' contributions} Study conception, design, development, analysis, preparation of all figures and tables, and main manuscript text: JMF. Section 3.3: BCG. Both authors interpreted, reviewed the text and results and approved the final version of the manuscript.

\FloatBarrier
\bibliography{sn-bibliography.bib}

\end{document}